\documentclass{article}
\usepackage{CJKutf8}
\usepackage{graphicx} 
\usepackage[T1]{fontenc}
\usepackage{ipa}
\usepackage[utf8]{inputenc}
\usepackage{amsmath}
\usepackage{appendix}
\usepackage{booktabs} 
\usepackage{float}
\usepackage{multicol}
\usepackage{multirow}
\graphicspath{{images/}}
\usepackage{xcolor}
\usepackage[export]{adjustbox}
\usepackage{natbib}
\usepackage{color}
\usepackage{tipa}
\usepackage{geometry}
\usepackage{hyperref}
\usepackage{comment}
\usepackage{authblk}
\usepackage{textcomp}

\title{A corpus-based investigation of pitch contours of monosyllabic words in conversational Taiwan Mandarin\thanks{Acknowledgments: This work was supported by the European Research Council under Grant SUBLIMINAL (\#101054902) awarded to R. Harald Baayen}}

\author[a]{Xiaoyun Jin}
\author[b]{Mirjam Ernestus}
\author[c]{R. Harald Baayen}

\affil[a]{Quantitative Linguistics, Eberhard Karls Universität Tübingen, 72074 Tübingen, Germany\\Email: xiaoyun.jin@uni-tuebingen.de}

\affil[b]{Center for Language Studies, Radboud University, 6525 HT Nijmegen, The Netherlands\\ Email: mirjam.ernestus@ru.nl}

\affil[c]{Quantitative Linguistics, Eberhard Karls Universität Tübingen, 72074 Tübingen, Germany\\ Email: harald.baayen@uni-tuebingen.de}

\date{12.09.2024}

\begin{document}
\begin{CJK}{UTF8}{bsmi} 
\begin{CJK}{UTF8}{gbsn} 
\maketitle

\pagestyle{plain}

\newpage
\begin{abstract}

\noindent
In Mandarin, the tonal contours of monosyllabic words produced in isolation or in careful speech are characterized by four lexical tones: a high-level tone (T1), a rising tone (T2), a dipping tone (T3) and a falling tone (T4). However, in spontaneous speech, the actual tonal realization of monosyllabic words can deviate significantly from these canonical tones due to intra-syllabic co-articulation and inter-syllabic co-articulation with adjacent tones. In addition, \citet{Chuang:Bell:Tseng:Baayen:2024} recently reported that the tonal contours of disyllabic Mandarin words with T2-T4 tone pattern are co-determined by their meanings. 

Following up on \citet{Chuang:Bell:Tseng:Baayen:2024} research, we present a corpus-based investigation of how the pitch contours of monosyllabic words are realized in spontaneous conversational Mandarin, focusing on the effects of contextual predictors on the one hand, and the way in words'  meanings co-determine pitch contours on the other hand.  We analyze the F0 contours of 3824 tokens of 63 different word types in a corpus of spontaneous conversational Taiwan Mandarin, using the generalized additive (mixed) model to decompose a given observed pitch contour into a set of component pitch contours.  These component pitch contours isolate the contributions to the pitch contour of the variables taken into account in the statistical model. We show that the tones immediately to the left and right of a word substantially modify a word's canonical tone.  Once the effect of tonal context is controlled for, the canonical rising (T2) and dipping (T3) tones emerge as low flat tones, contrasting with T1 as a high tone, and with T4 as a high-to-mid falling tone.  The neutral tone (T0), which in standard descriptions is taken to primarily depend for its realization on the preceding tone, emerges as a low tone in its own right, the realization of which is modified by the other predictors in the same way as the standard tones T1, T2, T3, and T4.  In line with the results from a previous study on disyllabic words with the T2-T4 tonal contour \citep{Chuang:Bell:Tseng:Baayen:2024}, we also show that word, and even more so, word sense, co-determine words' F0 contours, and that, as a consequence, heterographic homophones (e.g.,  的,  得, and  地) have their own tonal signatures.  Analyses of variable importance using random forests further supported the substantial effect of tonal context and an effect of word sense that is almost as important as that of  tonal context. \\

\noindent
\textbf{Keywords: tone, tone sandhi, sense-specific tonal realization, conversational speech, monosyllabic words, Taiwan Mandarin} 
\end{abstract}
\newpage

\section{Introduction} \label{sec:introduction}

The tone system of standard Mandarin Chinese is described as having four lexical tones (T1, high level; T2, rising; T3,  falling-rising; T4, falling) as well as a neutral tone \citep{chao1968grammar}, the realization of which is taken to depend on the preceding tone \citep{chao1930system,chien2021investigating}.  

When monosyllabic Mandarin words are produced carefully in laboratory speech, their tonal contours are similar to the canonical contours as described in the literature \citep{chao1968grammar,lai2008mandarin}. However, the actual tonal contours of tones in connected or spontaneous speech can differ substantially due to tonal co-articulation \citep{shen1990tonal,xu1997contextual}. \cite{xu1994production,xu1993contextual} found that, when the pitch values at the syllable boundaries are similar, such as for the tone sequence \textit{T4+T2+T4} (the low tail of the first falling tone fits the starting point of the T2, and the endpoint of this rising tone dovetails well with the starting point for the second falling tone), the production of T2 resembled the canonical T2 trend. However, when the end  of a tone does not well match the start of the next tone, then the tone contours of Beijing Mandarin in running speech may be substantially different.  

Not only inter-syllabic co-articulation but also intra-syllabic co-articulation may influence the realization of tonal contours. \cite{xu2003effects} analysed minimal pairs of Mandarin syllables in connected laboratory speech and found that the onset pitch of the vowel is higher when preceded by an unaspirated /t/ compared to an aspirated /t\super h/.  Furthermore, various cross-linguistic studies have provided evidence that different vowels may have their own intrinsic pitch properties \citep{ho1976acoustic,ladd1984vowel,bo1987vowel,whalen1995universality}. For instance, \cite{zee1980tone} found that in Taiwan Mandarin, the F0 of the vowel /u/ was the highest, followed by /i/, and then  /a/ and /\textschwa/. The experimental results of \cite{bo1987vowel} suggest, however, that, in standard Mandarin,  the intrinsic pitch of /a/ is not lower than /i/ and /u/ when produced with T3 and T4 tones by female speakers.

Recently, \citet{Chuang:Bell:Tseng:Baayen:2024} reported that the tonal pattern of Taiwan Mandarin disyllabic words with the T2-T4 tone pattern is partially determined by these words' meanings.  This result fits well with previous studies. \citet{plag2015homophony} observed that the duration of English word-final co-varies with the semantics that this segment is realizing, e.g., plural number on nouns, singular number on verbs, and possession. \citet{gahl_time_2024} show that the spoken word duration of English homophones is in part predictable from the meanings of these homophones.  The finding of \citet{Chuang:Bell:Tseng:Baayen:2024} has been replicated for words with the T2-T3 and T3-T3 tone patterns by \citet{lu2024form}. Considered jointly, these findings support the possibility that also in monosyllabic Mandarin words, the way in which pitch contours are realized is co-determined by semantics.  

However, it is at present unclear whether the results obtained by \citet{Chuang:Bell:Tseng:Baayen:2024} for disyllabic words with the T2-T4 tone pattern will generalize to monosyllabic words. Compared to disyllabic words, the meanings of monosyllabic words are substantially more ambiguous \citep{chen1992word,lin2010ambiguity}.  For example,  打 (d\v{a}) can be used in the sense of `to play' as in  打篮球 (d\v{a} l\'{a}n-qi\'{u}, play basketball), in the sense of `hit' as in  打桌子 (d\v{a} zhu\={o}-zi, `to hit the table'),  and in the sense of `to give a call' as in  打电话 (d\v{a} di\`{a}n-hu\`{a}). 

At the same time, monosyllabic words offer novel opportunities to test the hypothesis that words' meanings co-determine their pitch contours. Among disyllabic Mandarin words, heterographic homophones are scarce, and not frequent enough to be well-attested in small corpora of conversational speech.  However, among monosyllabic words, heterographic homophones, such as  弟 (d\`{\i}, `brother'), and  地 (d\`{\i}, 'ground'),  are more widespread.  If indeed words' meanings co-determine the realization of tone, then the prediction follows that the tonal realizations of heterographic homophones must be somewhat different.

The present study investigates the tonal realization of monosyllabic words with one of the vowels /a, i, u, \textschwa/, and carrying any of the different tones, in a corpus of spontaneous Taiwan Mandarin. In Mandarin, single-character words always correspond to single spoken syllables. The maximal rhyme structure of a syllable is CGVN: an onset consonant (C), a pre-nuclear glide (G), a vocalic nucleus (V) and a coda consonant (N), which, according to the phonotactic constraints on the Mandarin syllables, must be a nasal \citep{duanmu2007phonology}. Previous studies have found evidence for the merger of the two nasal codas /n/ and /\ng/, notably in southern dialects and also in Taiwan Mandarin \citep{chiu2019uncovering,chiu2021articulatory}. Further, final nasals are often not realized, or realized in the form of nasalization of the preceding vowel \citep{yang2010phonetic}. Whether the prevocalic glide is actually part of a diphthong is under debate \citep{duanmu2007phonology}.  In Mandarin, the vowel can be one of seven monophthongs as well as one of eight diphthongs \citep{yi1920lectures}. We restrict ourselves to the vowels /a, i, u, \textschwa/ and to syllables without coda consonants and glides to minimize segmental effects on the realization of the tones.

A first important goal of our study is to clarify how exactly in spoken conversational Taiwan Mandarin, words' pitch contours are co-determined by neighboring words and words' segmental properties.  One question of interest is to what extent tonal co-articulation modifies words' pitch contours.  One possibility is that these modifications are minor, and that the canonical tones remain clearly visible. Another possibility is that the effects of tonal co-articulation are pervasive, and profoundly change words' pitch contours.  If so, one would expect that only for words between pauses the canonical tone contours emerge clearly.  A further question of interest is  whether the articulatory mechanisms that have been put forward to account for tone sandhi in careful standard Mandarin speech \citep{chao1930system,xu1997contextual} are predictive for tone sandhi in spontaneous Taiwan Mandarin conversational speech?  A related question is whether the segmental effects on the realization of tone that have been reported for standard Mandarin \citep{ho1976acoustic,xu2003effects} are also detectable in conversational Taiwan Mandarin?

The second goal of the present study is to clarify, for spontaneous spoken Taiwan Mandarin, whether words' meanings co-determine their pitch contours, not only for bi-syllabic words with the T2-T4 pattern, but also for monosyllabic words with a wide range of canonical tones \citep[as described in][]{chao1930system,xu1997contextual,peng1997production} when other factors such as word duration or speech rate \citep{cheng2015mechanism,tang2020acoustic}, segmental properties \citep{ho1976acoustic,ladd1984vowel,bo1987vowel,whalen1995universality}, and tonal context \citep{shen1990tonal,xu1997contextual,xu1993contextual} are controlled for.   

The remainder of this paper is structured as follows. Section~\ref{sec:data} introduces the corpus that we used, how we extracted the pitch contours, the covariates that we took into consideration, and the method we used for statistical analysis.  Section~\ref{sec:results} reports the results obtained. The implications of our findings are laid out in the general discussion (section~\ref{sec:gendisc}).

\section{Method}\label{sec:data}

\subsection{Data}
\noindent
The corpus used in the current study is the Taiwan Mandarin spontaneous speech corpus \citep{fon2004preliminary}. Taiwan Mandarin is a variety of Mandarin with influences from southern Min \citep{norman1988chinese}. This corpus has been transcribed at the word level using traditional Chinese characters. We followed the transcriptions in the corpus, and distinguished between word types on the basis of the characters with which the words are transcribed.  In this corpus, there are 118,592 single-character word tokens, which represent 1471 orthographic word types. 

We used the Montreal Forced Aligner (henceforth MFA) \citep{mcauliffe2017montreal} to determine the boundary between the consonant and vowel in the words under study. In order to assess the accuracy of the results from the MFA, we randomly selected 10\% of aligned tokens and checked them manually. If the difference between the boundary from MFA and human transcriber was less than 20 ms, the result was considered as correct. The accuracy of the MFA was about 90\%, which we accepted as reasonable.

There are many word types and tokens with /a, i, u, \textschwa/ in the corpus: 53,139 tokens of 699 word types. About 85\% of the tokens belong to just 32 high frequency words, such as  的 (d\textschwa, genitive),  你 (n\v{\i}, you), and  他 (t\={a}, he).  In order to prevent model predictions from being heavily biased by the highest frequency words, we randomly sampled 200 tokens across 55 speakers for words with a token frequency greater than 200.  As a further measure to ensure the data are properly balanced, we also excluded those words with a token frequency lower than 10.

Subsequently, F0 was measured using the method described in \cite{wempe2018sound}, implemented in a script for Praat \citep{boersma19922022}. The script uses auto-correlation (without filtering) to determine pitch, with the pitch floor set at 75 Hz and the pitch ceiling at 500 Hz. Every 5 ms, the F0 value was extracted. Octave jumps were only found for slightly fewer than 1\% of the tokens.  We kept these tokens in the dataset. 

Unsurprisingly, for longer words, more pitch measurements were available. On average, each token had 14 measurement points. However, 14\% of the tokens had fewer than 5 measurement points,  which is problematic for statistical modeling. We therefore removed these tokens from our dataset.  The resulting dataset comprised 4024 tokens of 70 word types.

The meaning of every word token in the dataset was tagged using a word sense disambiguation system \citep{hsieh2024resolvingregularpolysemynamed} based on the Chinese WordNet \citep{huang2010constructing}. For the 70 words in our dataset, a total of 113 senses was identified. Most of the words have one to three senses. 70\% of the words have more than one meaning. Because sense is one of the main predictors in this study,  we ensured that all senses had sufficient tokens for model fitting. We therefore only included , for a given word, only those tokens whose senses were represented by more than 10 tokens. This left us with a dataset of 3824 tokens representing 101 senses across 63 word types.  

Table \ref{tab:vowel_exmaple} provides for each vowel under study  the number of types, the number of tokens, and an example word. 

\begin{table}[htpb]
    \centering
    \resizebox{0.6\linewidth}{!}{
    \vspace*{0.5\baselineskip}
\begin{tabular}{cccc} \hline
    Vowel & Types &  Tokens &  Mandarin Example  \\
    \hline
   i &  15 & 772 & \emph{ 你} (you, n\v{\i}) \\ \hline
   u &  11 & 493 & \emph{ 组} (group, z\v{u})\\ \hline
   \textschwa &  15 & 802 & \emph{ 和} (and, h\'{e})\\ \hline
   a &  22 & 1757 & \emph{ 打} (beat, d\v{a})  \\\hline
   
\end{tabular}
    }
    \caption{Overview of the datasets with the four vowels}
    \label{tab:vowel_exmaple}
\end{table}

\subsection{Predictors}

\noindent
In our analysis, we included the following predictors. 
\begin{description}

\item[Time] Time was normalised to the interval [0,1]. This normalization is essential for regression modeling with GAMs. Longer words had more measurement points in the [0,1] interval. 
\end{description}

\subsubsection{Speaker-related controls}
\begin{description}
\item[Gender] Gender serves as a main control variable, as female speakers tend to speak with a higher pitch and wider pitch ranges than male speakers \citep{gelfer2005relative,shen2011effect}.

\item[Speaker] To allow for differences in the average pitch height of individual speakers,  we included by-speaker random intercepts in our statistical models. 
\end{description}

\subsubsection{Context-related controls}
\begin{description}

\item[Tone sequence] Tones undergo co-articulation with adjacent tones \citep{shen1990tonal,mok2004effects}. To enable the GAM to account for the influence of neighboring tones on the realization of a word's pitch contour, we looked up in the corpus the pinyin of the preceding and following syllable, and extracted the canonical tones of these syllables from the pinyin.  There are 36 possible combinations of the preceding and following tone: each combination of the four lexical tones, one neutral tone, and NULL when the target word occurs next to a pause. All 36 possible combinations are attested in our dataset. As the effect of adjacent tones is expected to vary depending on the current tone, we created a factor, \texttt{tone sequence}, with as levels all attested pairings of 36 neighboring tones crossed with the 5 tones attested for our target words. Of the 180 possible combinations, 173 were attested in our dataset.  It is important to keep in mind that \textit{tone sequences} are established on the basis of the canonical tones of the target word and its immediate neighbors, and that \textit{tone sequence} does not take into account that the tones of the immediate neighbors themselves may be subject to co-articulation with their neighboring tones. 

\item[Utterance position] As the realisation of F0 contours in utterances is co-determined by sentence intonation \citep{ho1976acoustic,yuan2011perception}, we calculated the 0-1 normalised position of a target word in its utterance as covariate, following \citet{Chuang:Bell:Tseng:Baayen:2024}. It is difficult to differentiate, in spontaneous conversational speech, between utterance boundaries, hesitations, pauses, and interruptions by other speakers. In the present study, an utterance was defined as a sequence of words preceded and followed by a perceivable pause (regardless of its duration), following the annotations provided by the corpus. The normalised position of a given token in an utterance is the position at which the token occurs divided by the total number of words in the utterance. Hence, this predictor is bounded between 0 and 1. 

\item[Semantic relevance] Some studies \citep{hsieh2013prosodic,chen2006prosodic,ouyang2015prosody} found that both duration and F0 range of the tonal realizations of their target words were adjusted to the amount of information of the target word in the context. In this study, instead of using traditional measures of contextual predictability such as bigram probability, a new measure proposed by \citet{kun2024} was used: \textit{semantic relevance}.   This measure is the sum of the correlations of the contextualized embedding of a target word and the contextualized embeddings of the preceding three words as well as the following word.  This measure gauges how well a word's meaning fits with the meanings of the words in its immediate context.  We used \textit{semantic relevance} as a control variable. 
\end{description}

\subsubsection{Lexical predictors}
\begin{description}
\item[Duration] Previous studies have shown that the realization of pitch contours depends in part on word duration. \citep{howie1976acoustical,lin1989contextual,woo1969prosody,yang2017duration}. Mandarin monosyllables with T3 tend to be the longest and syllables with T4 the shortest  \citep{ho1976acoustic,whalen1992information}.  We included word \textit{duration} as a predictor.  We used log-transformed duration in order to avoid outlier effects for long word durations. Duration is strongly correlated with speech rate.  Including both \textit{duration} and speech rate gives rise to high collinearity and concurvity. As \texttt{duration} turned out to be the superior predictor, we included \texttt{duration} as predictor, and do not report further on speech rate. As the effect of \textit{duration} may play out differently over time, we used a tensor product interaction to allow for an interaction of \textit{time} and \textit{duration}.  It turned out that this interaction differed for female and male speakers.  

\item[Tone pattern]
We included \texttt{tone pattern} as a factorial predictor in our model. The \texttt{tone pattern} is the canonical tone of a word as specified in dictionaries and taught in textbooks \citep{xu1997contextual,peng1997production}. The levels of \texttt{tone pattern} are high (T1), rise (T2), dipping (T3), falling (T4), and neutral (T0) tone. Although in spontaneous speech, contextual factors such as tonal context, utterance position, and semantic relevance may alter the shape of the F0 contour of a monosyllabic word, it is generally assumed that these factors will not give rise to qualitative differences in the realization of the canonical tone patterns \citep{xu1997contextual,peng1997production}. In other words, the tonal contours in spontaneous speech are expected to resemble the canonical tone patterns to a considerable extent.  We therefore expect to find an independent contribution of \textit{tone pattern} to words' pitch contours. The reason that we included both \textit{tone pattern} and \textit{tone sequence} as predictors is to clarify whether an independent effect of \textit{tone pattern} is present over and above the effect of \textit{tone sequence}. In other words, do word tokens that share the same tone pattern but that occur in different tonal contexts have similar pitch contours?

\item[Consonant]
An effect of onset consonant on pitch has been reported in various studies of Mandarin phonetics \citep{xu2003effects,ho1976acoustic}. \cite{xu2003effects} found that the onset F0 of a tone is higher following unaspirated consonant /t/ than following its aspirated counterpart /t\super h/.  As we expect that not only aspiration but also other consonantal properties may influence the realization of pitch contours, we included \textit{consonant} as factorial predictor. In our dataset, 19 different consonants are attested. Three word types do not have onsets. These words were coded for having a \textsc{null} onset. The factor \texttt{consonant} therefore has 20 levels in our dataset. 

When \textit{consonant} is included as a predictor in order to take into account the mechanical effects of consonant articulation on pitch contours, it has as simultaneous effect that, compared to \textit{tone pattern}, the set of possible words is narrowed down substantially.  (Note, however, that even in combination with the vowel and information about the tone, the identity of a word is often still not fully determined due to the extensive homophony in the Mandarin lexicon.)   Potential effects of \textit{consonant} thus have two possible sources: the physics of articulation on the one hand, and the word and its semantics on the other hand.

\item[Word]
The factorial predictor word has as levels the characters of the individual words  with which they are represented in the orthographic transcription of the Taiwan Mandarin corpus. Our dataset contains 6 sets of heterographic homophones: four doublets (ni3, ge4, di4, and bu4), a triplet (de0), and a quadruplet (ta1).  The only character in our dataset that has more than one possible segmental realization is  地, which can be pronounced either as d{e}0 or d\`{i}. In addition, a few characters have more than one possible tonal realization, and hence more than one meaning (e.g.,  哪, na0/n\v{a}, exclamatory tone/where)). 

\item[Word sense]\label{sec:word sense}
Monosyllabic words in Mandarin often have many different meanings or senses\citep{chung2006mandarin}. In our dataset of 3824 tokens across 63 word types, \textit{sense} is a factorial predictor with 101 levels. Some senses are shared by multiple words with different pronunciations. For example, the sense `to explain or remind the other' is shared by the sentence-final particles  啊 (a0) and  哪 (na0). It is conceivable that the senses of words that are not pronounced the same are not fully identical. We therefore created a second factor, \textit{word sense}, with 110 levels, which assigns different meanings to different word types (e.g., also to  啊 (a0) and  哪 (na0)). This affected five sentence-final particles with the vowel /a/ and two words with /\textschwa/. 
\end{description}

\subsection{Statistical modeling}

\noindent
F0 contours are non-linear functions over time. We used the generalized additive model (GAM) to predict tone contours from the above-mentioned predictors, using  the \textbf{mgvc} package  \citep{wood2015package} in \citet{r2013r}.  We log-transformed the F0 values before analysing them as our response variable. We fitted separate models for each of the four vowels to avoid the necessity to have each predictor interact with vowel identity. However, we also report one omnibus model for the data of all four vowels jointly,  including the best performing predictors.

Pitch contours are time series with slowly changing values that are inevitably auto-correlated.  One common strategy to deal with auto-correlation in GAMs is to incorporate an AR(1) process.  Inclusion of an AR(1) process with auto-correlation of about $\rho  = 0.9$ removed nearly all auto-correlation from the residuals.  Unfortunately, the resulting distributions of the residuals of Gaussian models fitted to the pitch contours have heavy tails, which resist correction by assuming the residuals are following a t-distribution \citep[see also][]{Chuang:Bell:Tseng:Baayen:2024}.  As most GAMs including both an AR(1) process and the scaled t-distribution did not converge properly, the models reported below did not correct for the distribution of the residuals, but did include an AR(1) process.  Including the AR(1) process resulted in substantially more precise fits of the model to the data. 

For each of the four vowels, we started out with a baseline model with all control predictors, and one additional core predictor, denoted by $X$, which was either \textit{tone pattern}, \textit{consonant},  \textit{word}, or \textit{word sense}. These four predictors were not simultaneously entered in the model because they are too collinear.  (In section~\ref{sec: gam_randomforest}, we report a random forest analysis that takes all predictors into account simultaneously.) An additional advantage of  considering separate models, apart from avoiding collinearity and concurvity, is that it becomes straightforward to assess the relative importance of these four variables using Akaike's information criterion \citep[AIC,][]{Akaike:1974}. 
\newpage
\begin{tabbing}
mm \= \texttt{pitch(logF0)} \= $\sim$ \= \texttt{gender} + \kill
    \> \texttt{pitch(logF0)} \> $\sim$ \> \texttt{gender} +  \\
    \> \> s(\texttt{time}, by=gender, k = 4) + \\
    \> \> s(\texttt{duration}, by=gender, k = 4) + \\
    \> \> ti(\texttt{time}, \texttt{duration}, k = c (4, 4)) + \\
    \> \> s(\texttt{speaker}, bs = ``re") + \\
    \> \> s(\texttt{time}, \texttt{tone sequence}, bs = ``fs", m = 1) + \\
    \> \> s(\texttt{utterance position}, k = 4) + \\
    \> \> s(\texttt{semantic relevance}, k = 4) + \\
    \> \> \texttt{X} + s(\texttt{time}, by = \texttt{X}, id = 1, m = 1) \\
\end{tabbing}

The first 8 terms of this model specification control for \textit{gender}, normalised \textit{time}, log-transformed \textit{duration}, \textit{speaker}, \textit{tone sequence}, \textit{utterance position}, and \textit{semantic relevance}.  We constrained the number of basis functions for \texttt{time}, \texttt{duration}, the interaction between \texttt{time} and \texttt{duration}, \texttt{utterance position} and \texttt{semantic relevance} to 4 to avoid overfitting.  The last line of the model specification requests an intercept and a separate smooth for \textit{time} for each level of \texttt{X}. These smooths are random effect smooths \citep[see][for detailed discussion]{baayen2022note}, the non-linear equivalent of combined random intercepts and random slopes in linear mixed models. 
  
\section{Results}\label{sec:results}

\noindent
In this section, we first present the effects of the control variables. These effects are mostly independent of which predictor \texttt{X} is included, and are therefore reported for a model that does not include \texttt{X}, henceforth our \textit{baseline model}.  We subsequently report the central analyses using the four core predictors \texttt{tone pattern}, \texttt{consonant}, \texttt{word}, and \texttt{word sense}.  This section concludes with an omnibus analysis of all vowels jointly, complemented with a random forest analysis.

\subsection{Control variables}

\subsubsection{Gender and duration by gender}
\noindent
As expected, on average, the pitch contours of female speakers are higher than the pitch contours of male speakers. For the full dataset with all vowels, log F0 was on average 0.41 log units lower for males (see Table~\ref{tab.gam}).

\begin{figure}[H]
    \centering
    \includegraphics[scale=0.5]{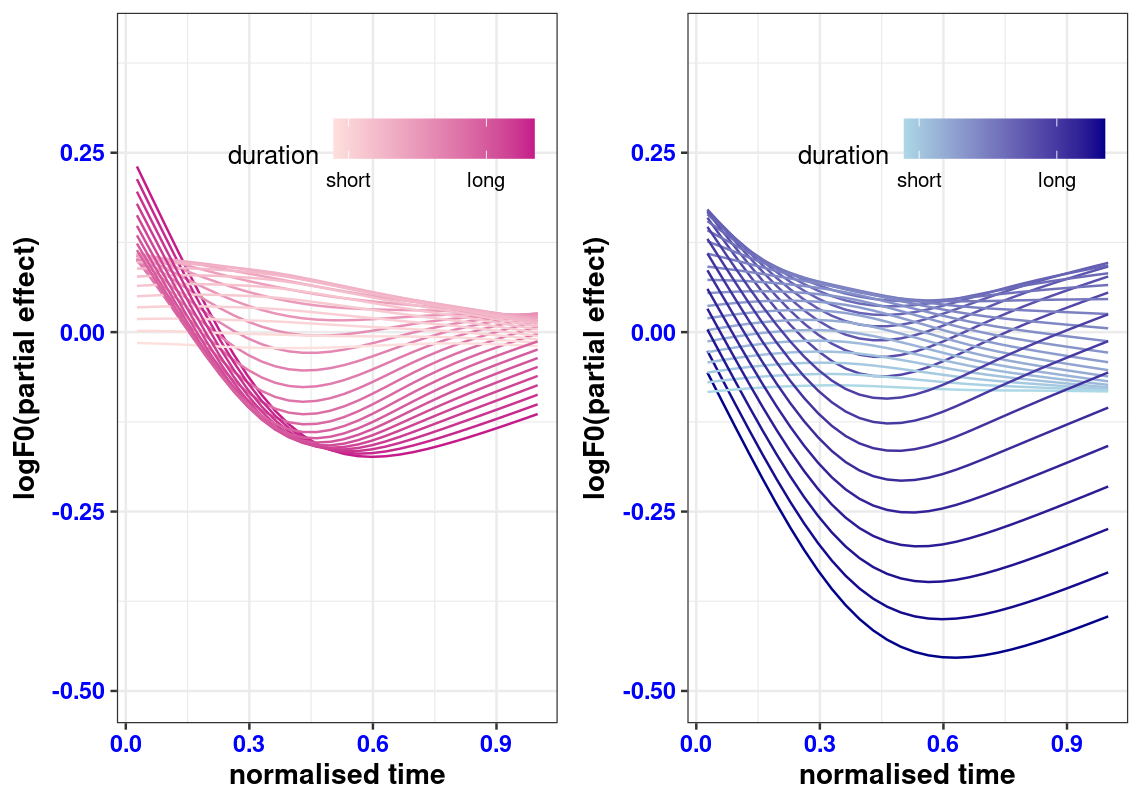}
    \caption{The partial effect of log-transformed \textit{word duration} by time for female (left) and  male speakers (right), on the pitch contour for words with /u/. Darker colors indicate longer word durations.}
    \label{figure:duration}
\end{figure}

The effect of \textit{duration} on pitch was different for the two sets of speakers, as illustrated in Figure~\ref{figure:duration} for words with /u/ as vowel. For shorter words  (lighter colors), the partial effect of \textit{duration} on the pitch contours is almost flat and close to zero, for both genders.  For longer word tokens, the partial effect for females first falls,  then rises, and finally levels off.  For male speakers, the partial effect on the pitch contour resembles that of female speakers, but with a broader range.  Appendix~\ref{appendix duration} provides further information on the partial effect of the interaction of \texttt{time} by \texttt{duration} by \texttt{gender} for the other three vowels.

\subsubsection{Speaker}

\noindent
The random intercepts for speaker were well-supported across all GAM models. They enable further fine-tuning of the difference in average pitch height within the two genders.  We explored more complex GAMs in which speakers have their own pitch contour signatures, modeled with by-speaker factor smooths for \textit{time}, but this did not improve model fit,  while requiring substantially increased computation time. We therefore only included by-subject random intercepts.

\subsubsection{Tone sequence}\label{sec:tone sequence}
\begin{figure}[htbp]
    \centering
    \includegraphics[width=1\textwidth]{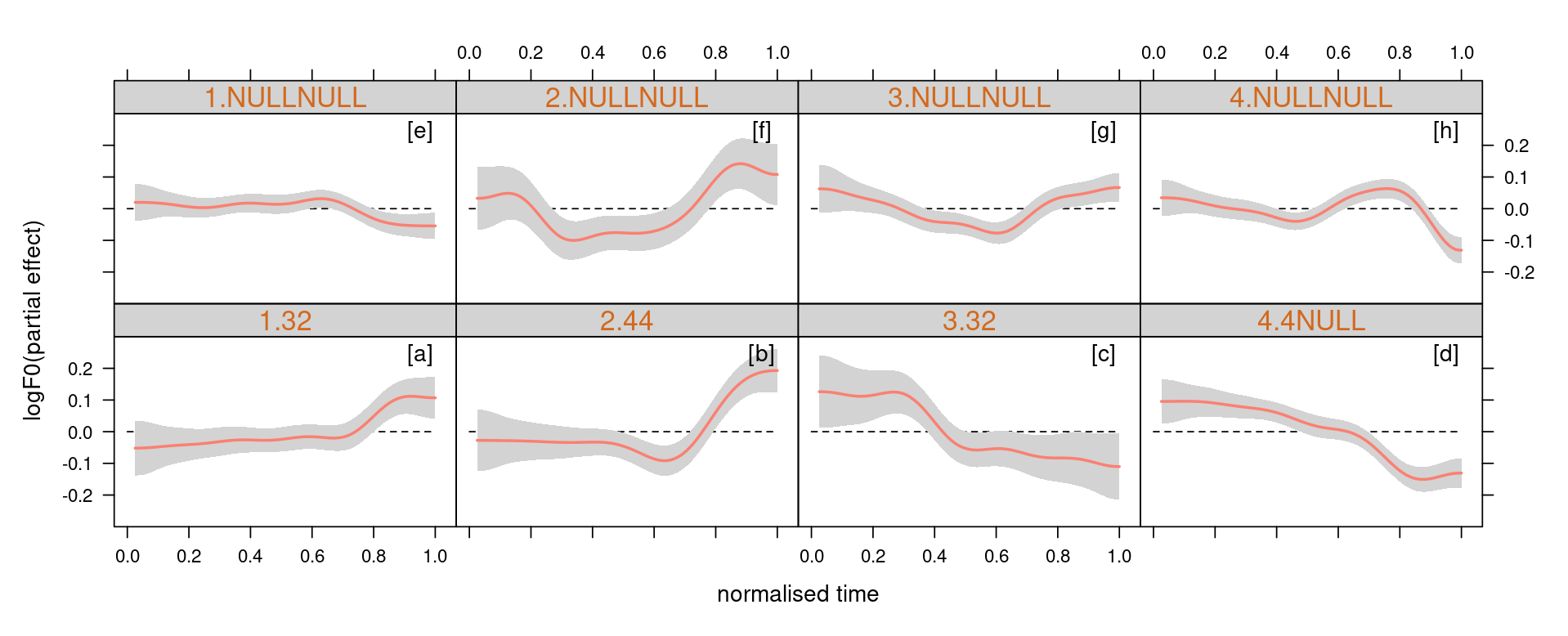}
    \caption{Partial effect of \texttt{time} by \texttt{tone sequence}, for words, with /a/,  preceded and followed by pauses (upper panels), and for /a/ words with selected combinations of preceding and following tones (lower panels). The first number of a panel legend denotes the tone pattern of the token itself, the numbers following the dot specify the preceding and following tones respectively.  If there is no preceding or following tone, this is coded as `NULL'. }
    \label{fig:adj-tone}
\end{figure}

\begin{figure}[H]
    \centering
    \includegraphics[width=1\textwidth]{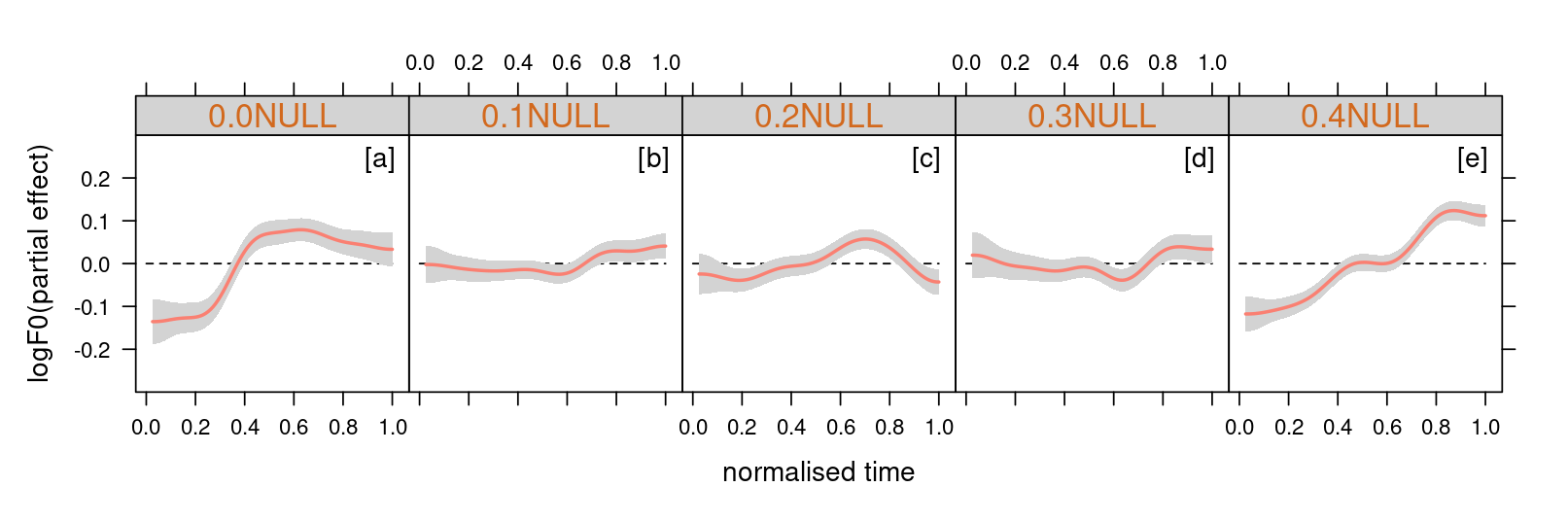}
    \caption{Partial effect of \texttt{time} by \texttt{tone sequence}, for words with /a/ that carry a neutral tone and that are preceded by one of the five tones and that are followed by a pause (coded as `NULL').}
    \label{fig:adj-tone:neutral}
\end{figure}

The partial effect of \textit{tone sequence} on the pitch contours is illustrated in Figure~\ref{fig:adj-tone} and Figure~\ref{fig:adj-tone:neutral} respectively. These examples all pertain to words with /a/, the only vowel for which all five tones (the four canonical tones and the neutral tone) are attested in our dataset.  The panels in the top row (e--h) of Figure~\ref{fig:adj-tone} illustrate the partial effect of tone sequence on tonal realization for words between pauses.  

Recall that \textit{tone sequence} is nested under \textit{tone pattern}. If \textit{tone pattern} has a main effect that is much stronger than that of \textit{tone sequence}, then the expectation is that the effect of \textit{tone sequence} for words between pauses is minimal, or even absent.  In this case, level contours with wide confidence intervals are predicted for the partial effect for \textit{tone sequence}.  However, if the main effect of \textit{tone pattern} is small compared to the effect of \textit{tone sequence}, then one might expect the canonical tones to be most clearly visible for words between pauses.  Below, we document that the main effect of \textit{tone pattern} is very small. As a consequence, for words between pauses, we expected the canonical tone pattern to emerge.  However, this expectation is only partially supported by the data. In the absence of adjacent tones (coded as \textsc{nullnull}), we see a level tone with a slight final dip for T1, an early fall followed by a long rise for T2, a dipping tone for T3, and a late rise-fall for T4.  The only tone that is realized with a contour that is similar to the expected canonical tonal contour is the dipping tone (T3).

The bottom panels (a-d) of Figure~\ref{fig:adj-tone} illustrate the effects of neighboring tones.  A word of caution is in order here. The neighboring tones are the canonical tones of the preceding and following syllables. However, as mentioned above, these tones themselves are subject to tonal assimilation. As a consequence, the descriptions of the tone sandhi patterns presented in what follows can only be approximate, as the actual realization of the adjacent tones is not taken into account.

The contour in panel (a) shows a high tone (T1) preceded by a dipping tone (T3) and followed by a rising tone (T2). The pitch contour is flat, except for a final rise that might be anticipating the rise of the following T2.

Panel (b) of Figure~\ref{fig:adj-tone} illustrates that a rising tone (T2) surrounded by falling tones (T4) starts out stationary, and then rises continuously. This finding is partly in line with  \cite{xu1994production}, who found that a rising tone in this tonal environment is continuously rising. In our dataset of Taiwan Mandarin, however, only the final half of the contour is rising, while the first half is flat.

\citet{chao1948mandarin} observed that when a dipping  tone (T3) is followed by another dipping tone, the first one is pronounced as a rising tone (i.e., as T2).  Although for standard Mandarin this tone sandhi has been reported to be incomplete \citep{cui2020effect}, in conversational Taiwan Mandarin, sequences of 2-3 and 3-3 tones are indistinguishable \citep{lu2024form}. The starting point of the tone contour in panel c, showing the realization of a dipping tone after another dipping tone and before a rising tone  (i.e., in the sequence 3-3-2), is therefore as expected. The fact that the second half of the contour is flat indicates that, in our dataset, the `dipping tone' only has an initial fall, and no subsequent rise.  The final rise may have been neutralized by the following rising tone (T2). 

According to \citet{chao1948mandarin} and \citet{shen1990mandarin}, when T4 is followed by a second T4, the first T4 does not fall quite to the bottom, but only to the middle.  This may explain why, when T4 is preceded by T4 and followed by a pause (panel d), it starts out somewhat higher compared to when it is realized in isolation.  However, why a clear fall is visible in panel d, whereas in panel h we find a rise before the final fall, remains unexplained. 

When it comes to the neutral tone, the literature describes its realization as depending on the preceding tone \citep{chao1930system,yip1980tonal,shen1990tonal,chien2021investigating}. Following a dipping tone (T3), a neutral tone would become a rising tone. However, in Figure~\ref{fig:adj-tone:neutral}, the pitch contour of a neutral tone preceded by a T3 (panel d)  is mostly flat. Furthermore, the neutral tone has been argued to start with the value of the end of the preceding tone and then to be realized with a fall after T1, T2 and T4 \citep{yip1980tonal,shen1990tonal}.  However, in our data, the neutral tone contour after T1 (in panel b) is basically flat with possibly a minute rise near the end, instead of having the expected fall.  When preceded by T2 (panel c), the neutral tone shows a very modest rise, followed by the expected fall. For a neutral tone after T4 (panel e), we observe a rising contour without a final fall.  When preceded by another neutral tone, the second neutral tone shows a rise and then levels off. These observations lead to the conclusion that, in Taiwan Mandarin, the neutral tone is not realized in the same way as in standard Mandarin, which is consistent with the findings of \cite{tseng2004prosodic}, who concluded that  in Taiwan Mandarin the neutral tone is independent from the preceding tone.

\subsubsection{Utterance position}

The baseline model requests a main effect smooth for \textit{utterance position}. The left panel of Figure~\ref{sm_F0)} shows how the tonal shapes of the four vowels are influenced by their position in the utterance.  The more a word occurs towards the end of an utterance, the lower its F0. This is consistent with the finding in \cite{shih1997declination} that statement intonation is often characterised by a downward inclination. 

\subsubsection{Semantic relevance}

\begin{figure}
\begin{minipage}[]{0.48\textwidth}
\centering
\includegraphics[width=1.0\textwidth]{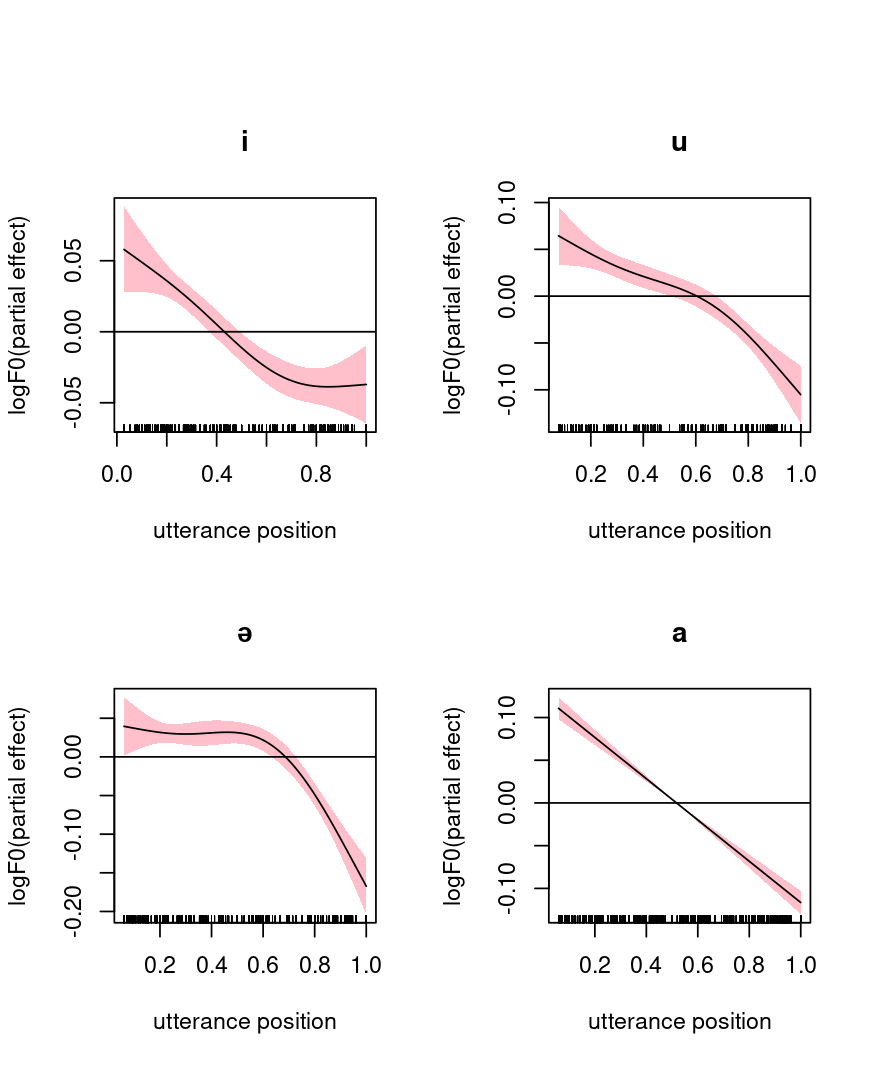}
\end{minipage}
\centering
\begin{minipage}[]{0.48\textwidth}
\centering
\includegraphics[width=1.0\textwidth]{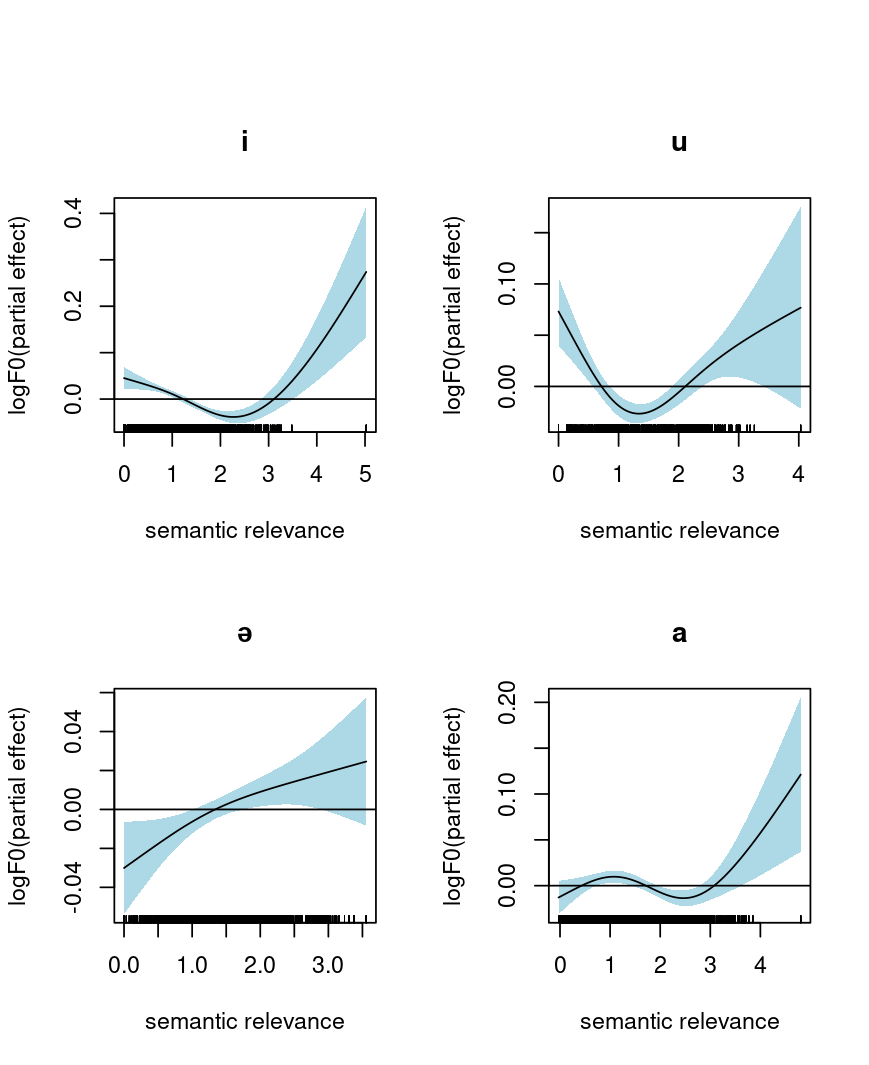}
\end{minipage}
\caption{Left panel: the partial effect of \textit{utterance position}; Right panel: the partial effect of \textit{semantic relevance}. Note that the range of the log-transformed pitch on the y-axis differs across the vowels.}
\label{sm_F0)}
\end{figure}

The right panel of Figure~\ref{sm_F0)} presents the partial effect of \textit{semantic relevance}. As there are only a few data points with large values of \textit{semantic relevance}, confidence intervals for these large values are very wide. For words with /i/ and /u/, the partial effect of \textit{semantic relevance} is similar and U-shaped. For words with  /\textschwa/, F0 first increases with increasing \textit{semantic relevance}, and then levels off.  Semantic relevance does not have a clear effect for words with /a/.

\subsection{Tone pattern} \label{sec:canonical tone}

\begin{figure}[H]
    \centering
    \includegraphics[width=1\textwidth]{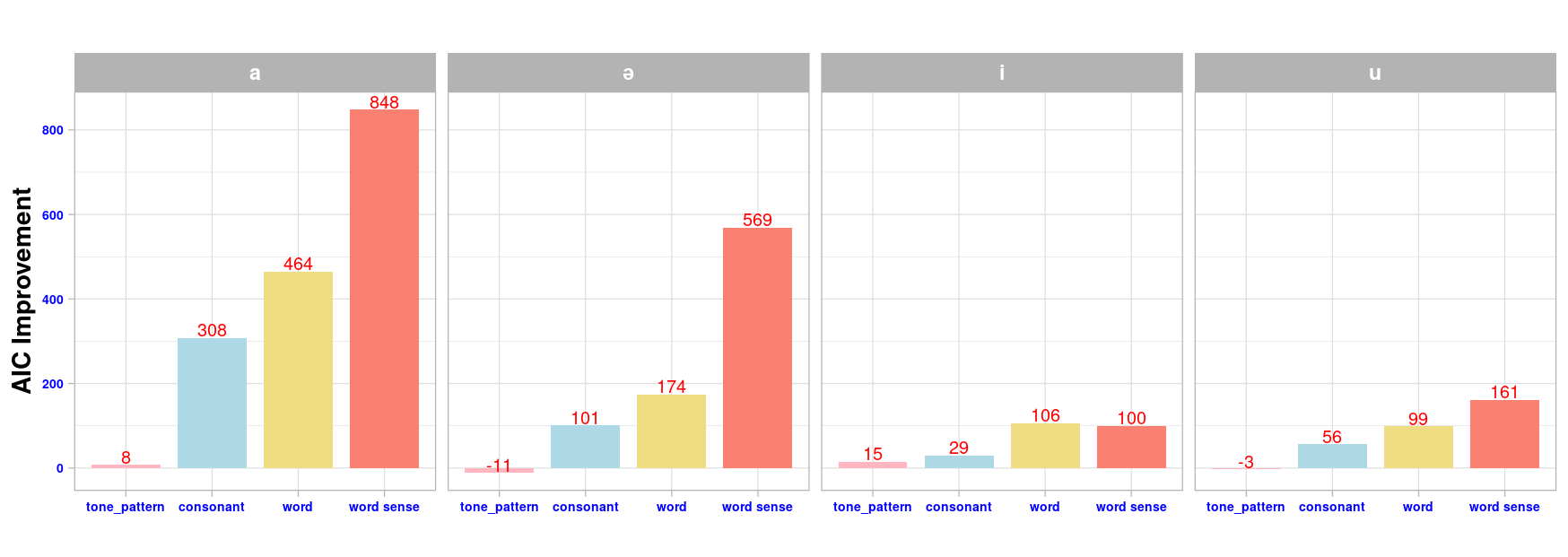}
    \caption{Improvement in model fit (using AIC), for each of the four vowels, when adding \textit{tone pattern} (pink bars), \textit{consonant} (blue bars), \textit{word} (yellow bars) and \textit{word sense} (red bars) as predictor to the baseline model.}
    \label{fig:AIC_F0}
\end{figure}

\begin{figure}[H]
    \centering
    \includegraphics[width=1\textwidth]{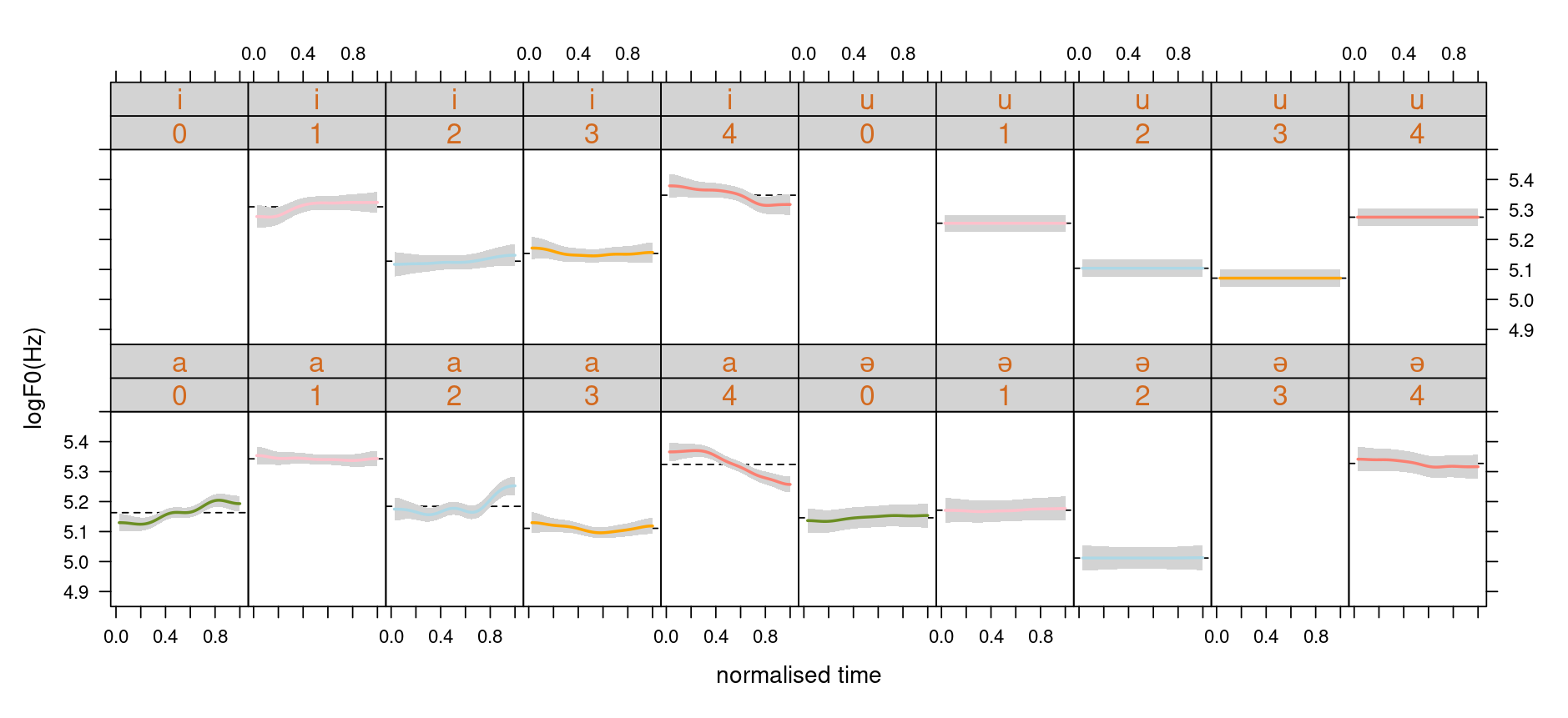}
    \caption{Partial effect of \textit{tone pattern} in interaction with normalised \textit{time}, for the vowels /i, u, a, \textschwa/. The five different tone patterns are highlighted by different colours. Dashed lines denote the intercept for each tonal pattern predicted by the GAM.  Empty panels pertain to vowel-tone combinations for which we have no data.}
    \label{fig:tone pattern}
\end{figure}

Surprisingly, \textit{tone pattern} hardly improved model fit (see the pink bars in Figure~\ref{fig:AIC_F0}). Reductions in AIC were modest (8 for /a/, 15 for /i/) or even negative (-11 for /\textschwa/, -3 for /u/).   The partial effect of \textit{tone pattern} is almost always invariant over normalised \textit{time}.  As shown in Figure~\ref{fig:tone pattern}, the only potential exception is the falling tone (T4) when realized on words with /i/ or /a/.  Across tone patterns and vowels, confidence intervals are wide and continuously include the tone-specific intercepts, indicating that the tone patterns are basically level tones. In fact, the effect of the canonical \textit{tone pattern} in the statistical models is present mainly in the form of changes to the intercept, that is, what we have is clear evidence for differences in the overall pitch height of the tones.  Figure~\ref{fig:tone pattern} suggests that, for Taiwan Mandarin, most tone patterns are realized as level mid tones.  High level tones are observed for /i/ for T1 and T4, for /u/ for T1, for /a/ for T1 and T4, and for /\textschwa/ for T4. Low level tones are present for /u/ and /\textschwa/ for T2.

\begin{figure}[H]
\begin{minipage}[]{0.48\textwidth}
\centering
\includegraphics[width=1.0\textwidth]{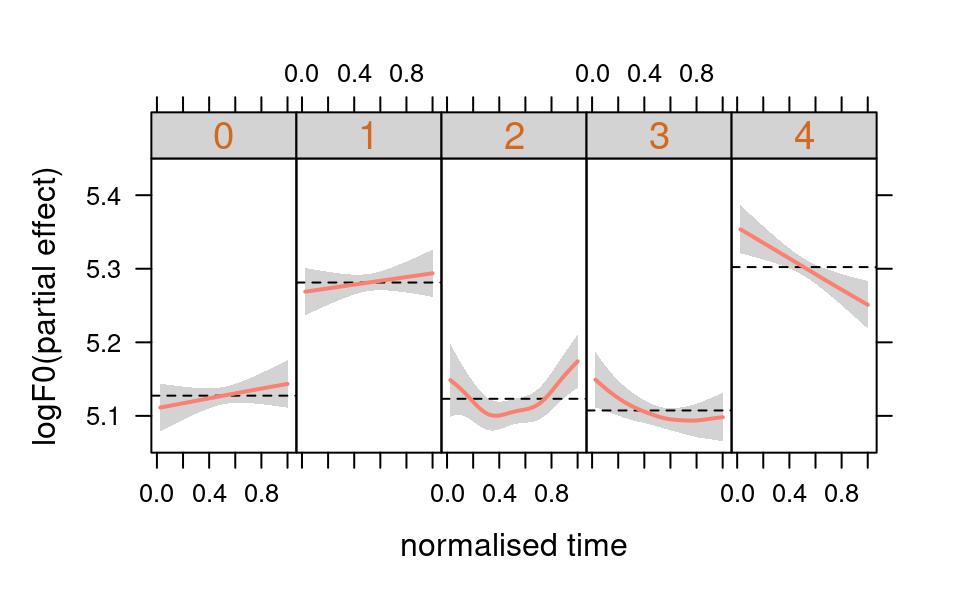}
\end{minipage}
\centering
\begin{minipage}[]{0.48\textwidth}
\centering
\includegraphics[width=1.0\textwidth]{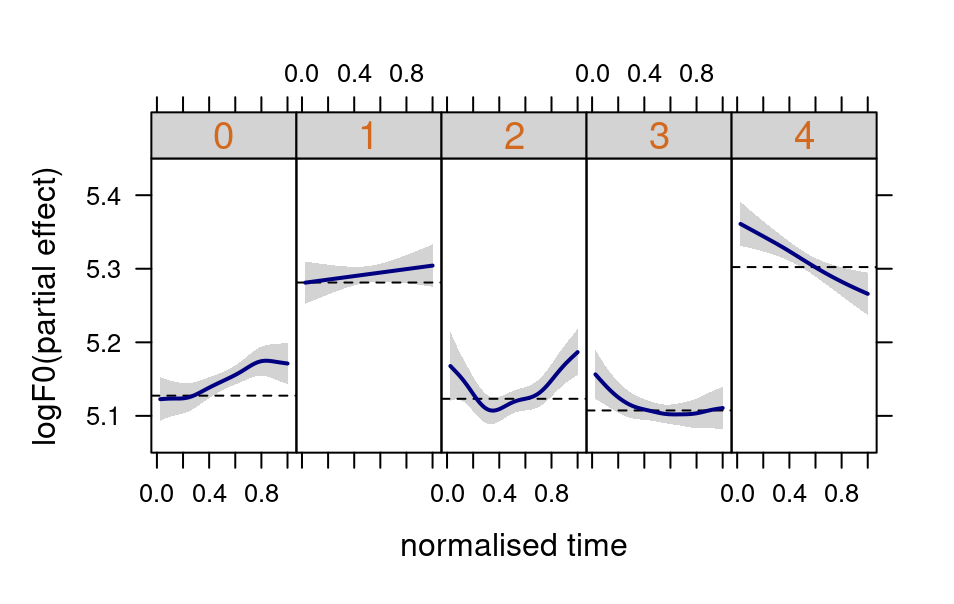}
\end{minipage}
\caption{Partial effect of \textit{tone pattern} for two omnibus GAM models (fitted to the data of all four vowels jointly). The left panel presents the partial effects for a GAM including \textit{tone sequence}, the right panel presents the corresponding partial effects for a GAM without this control predictor.  Removal of \textit{tone sequence} leads to a slight overestimation of non-linearity for T0, T2 and T3.}
\label{fig:tone_patterns_means}
\end{figure}

\textit{Tone sequence} is nested under \textit{tone pattern}, just as \textit{speaker} is nested under \textit{gender}. In order to clarify to what extent the effect of \textit{tone pattern} depends on \textit{tone sequence}, we fitted two omnibus models to the pitch contours of the combined data of all four vowels, including a main effect for vowel.  The first model includes \textit{tone sequence} as control variable, the second model does not.    The partial effects of \textit{tone pattern} of these omnibus GAMs (models not shown) are presented in Figure~\ref{fig:tone_patterns_means}. The partial effects of \textit{tone pattern} with \textit{tone sequence} as control are shown in the left panel. This panel confirms that the main differences between the tones concern pitch height: T0, T2 and T3 are low tones, T1 is a high tone, and T4 has a modest fall.  When \textit{tone sequence} is removed as control predictor, the  general effect of \textit{tone pattern} remains very similar, as shown in the right panel of Figure~\ref{fig:tone_patterns_means}.  In this simplified model, in which the effect of neighboring tones on the realization of pitch contours is ignored, there is slightly more support for T0 being a rising tone, and T2 having a (late) rise.  However, inclusion of \textit{tone sequence} as control predictor clarifies that the small contour effects for T0 and T2 that are visible in the model without \textit{tone sequence} are spurious:  \textit{tone sequence} is a much more critical predictor than \textit{tone pattern} itself (the AIC of the model with \textit{tone sequence} is 2298.48 units lower than the AIC of the model without \textit{tone sequence}, whereas the reduction in AIC when \textit{tone pattern} is removed from the model while keeping \textit{tone sequence} as predictor is only 22.46 units).

We have seen that the contribution of \textit{tone pattern}, a lexical variable, to model fit is small.  The study by \citet[][]{Chuang:Bell:Tseng:Baayen:2024} suggests the possibility that word might be a more important lexical predictor than \textit{tone pattern}, independently of \textit{tone sequence}.  As a first step towards demonstrating the importance of word and word meaning, we consider as predictor the initial \textit{consonant} (coded as `NULL' when absent) as a lexical predictor: given knowledge about the initial consonant, uncertainty about word identity is considerably reduced compared to just having information about the tone pattern. Furthermore, this analysis will also be informative about the extent that the articulatory properties of consonants \citep{xu2003effects,ho1976acoustic} co-determine the realization of words' F0 contours.

\subsection{Consonant-specific tonal contours}

\begin{figure}[H]
    \centering
    \includegraphics[width=1\linewidth]{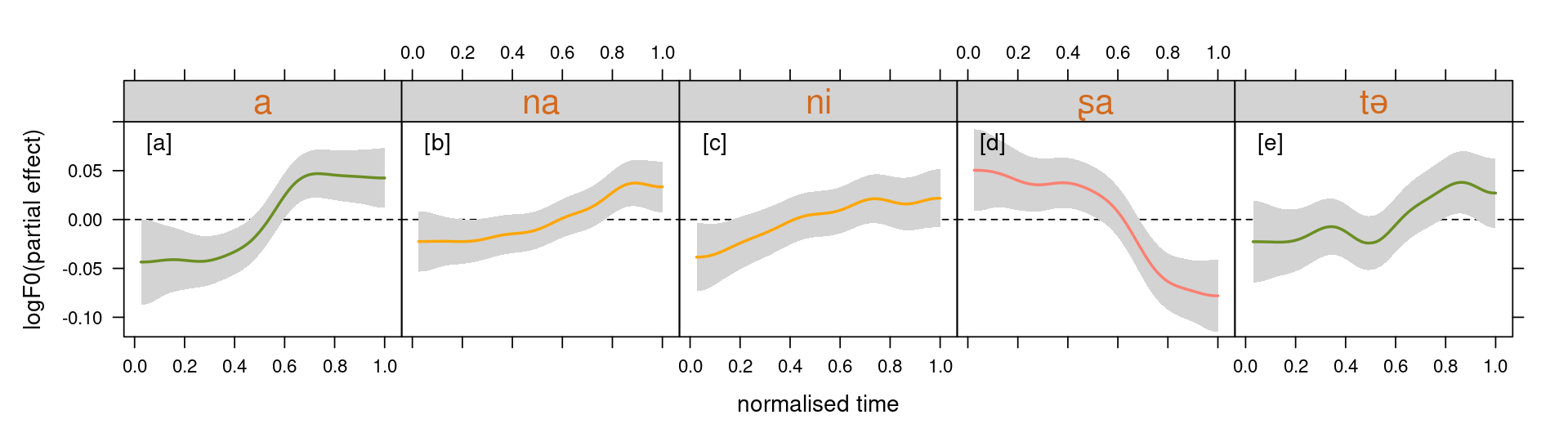}
    \caption{Partial effect of \textit{consonant} in interaction with normalised \textit{time}. Only the consonants for which the partial effect is significant are included. The consonants and vowels are marked in the title of each panel, while identical consonants also share the color of their lines (na and ni).}
    \label{fig:consonant}
\end{figure}

We added by-consonant random smooths for normalized \textit{time} to the baseline GAM models fitted to the four vowel-specific datasets.  The blue bars in Figure~\ref{fig:AIC_F0} clarify that the contributions to the model fits are substantially larger for \textit{consonant} than for \textit{tone pattern}, across all four vowels.    Panel (a) of Figure~\ref{fig:consonant} shows the contour for words without a consonant. Here the main trend is a rise that is constrained to the middle of the vowel. The partial effects for the three consonants with clear effects are presented in the remaining panels of Figure~\ref{fig:consonant}. For words with an initial /n/ (na/ni in panel b and c), a modest slow rise can be observed.  For words with an initial /\textrtails/ (panel d), a downward pitch movement emerges.  For the word pronounced as /t\textschwa/, a rising contour is observed.

\begin{figure}[H]
    \centering
    \includegraphics[width=1\linewidth]{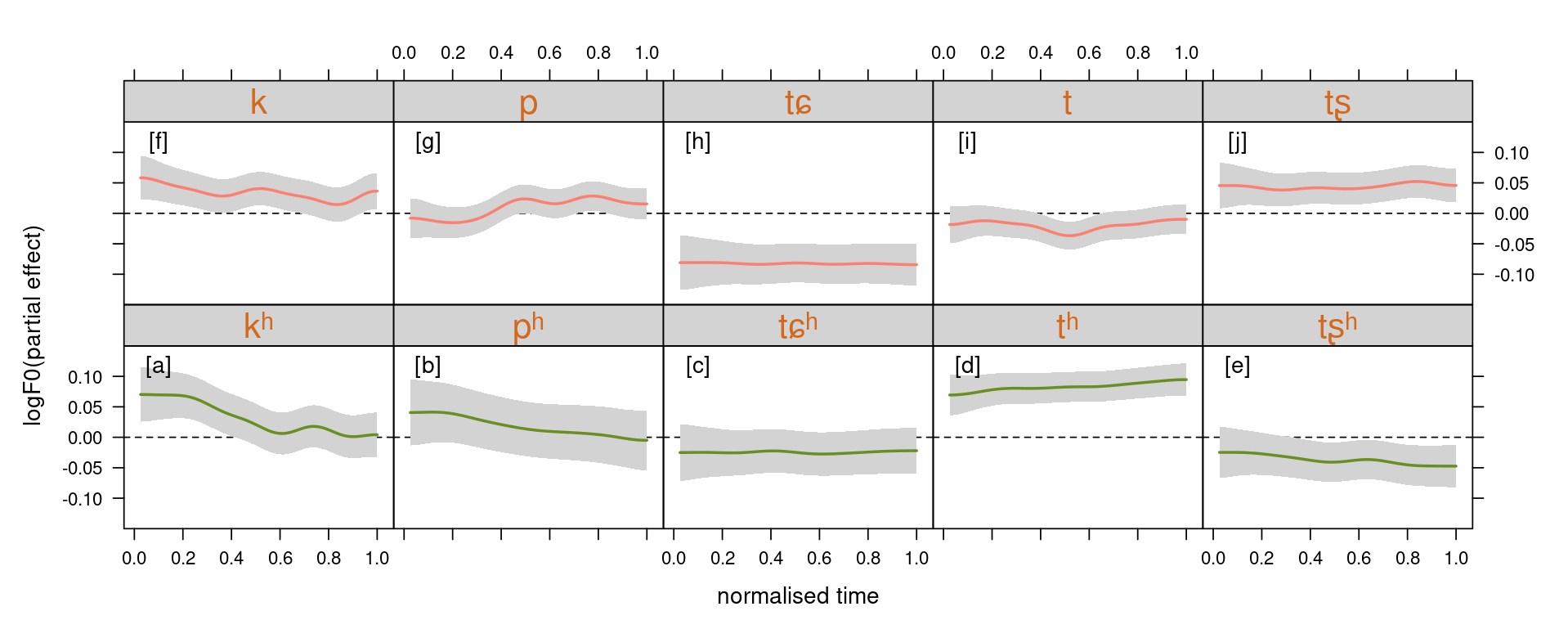}
    \caption{Partial effect of \textit{consonant} in interaction with normalised \textit{time}.}
    \label{fig:consonant_aspiration}
\end{figure}

\cite{xu2003effects} observed for laboratory speech that the onset F0 of a tone is higher for words with an unaspirated /t/ than for words with an aspirated /t\super h/, with /a/ as vowel. In the present study, five pairs of aspirated and unaspirated consonants are attested across four vowels. Therefore, we fitted an omnibus GAM model fitted to the data of all four vowels jointly, with by-consonant random smooths (model not shown). The partial effect of the five unaspirated consonants /k, p, t\textctc, t, t\textrtails/ and their corresponding aspirated ones are presented in Figure~\ref{fig:consonant_aspiration}. For /t\super h/ and /t/ (panel d and i), we see a higher positioned partial effect for the aspirated consonant, exactly the opposite of what \cite{xu2003effects} reported for /ta/ and /t\super ha/. A similar effect of aspiration is present for words with /t\textctc \super h/ and /t\textctc/ (panels c and h).  For the other three pairs of aspirated and unaspirated consonants, differences are negligible (panels b and g, and a and f), or go in the opposite direction (panels e and j), showing a pitch onset that is lower for unaspirated consonants instead of being higher.

These results lead to the conclusion that, at least for conversational Taiwan Mandarin, there is no reliable evidence that aspiration necessarily leads to lower pitch.  Different consonants are sometimes associated with somewhat different F0 contours.  Having determined that their precise effects are not straightforward to link to articulatory constraints, we next consider the possibility that the effects we see are, at least in part, confounded with a lexical effect rather than a segmental effect, namely, the effect of word.

\subsection{Word-specific tonal contours}

\noindent 
Following up on the study by \citet{Chuang:Bell:Tseng:Baayen:2024}, we now consider a GAM that includes a non-linear random effect of word by normalised \textit{time}, using a factor smooth.  As our dataset comprises 64 words, this resulted in 64 word-specific partial effect pitch contours. The AIC scores (yellow bars) shown in  Figure~\ref{fig:AIC_F0} indicate that across all four vowels, models with a by-word factor smooth improve substantially on the models with \textit{consonant} or \textit{tone pattern}, replicating for monosyllabic words the observation made by \citet{Chuang:Bell:Tseng:Baayen:2024} for two-syllable words with the T2-T4 tone pattern, and a similar observation made by \citet{lu2024form} for two-syllable words with the T2-T3 and T3-T3 tone patterns.

\begin{figure}
    \begin{minipage}[]{0.48\textwidth}
        \centering
        \includegraphics[width=1\textwidth]{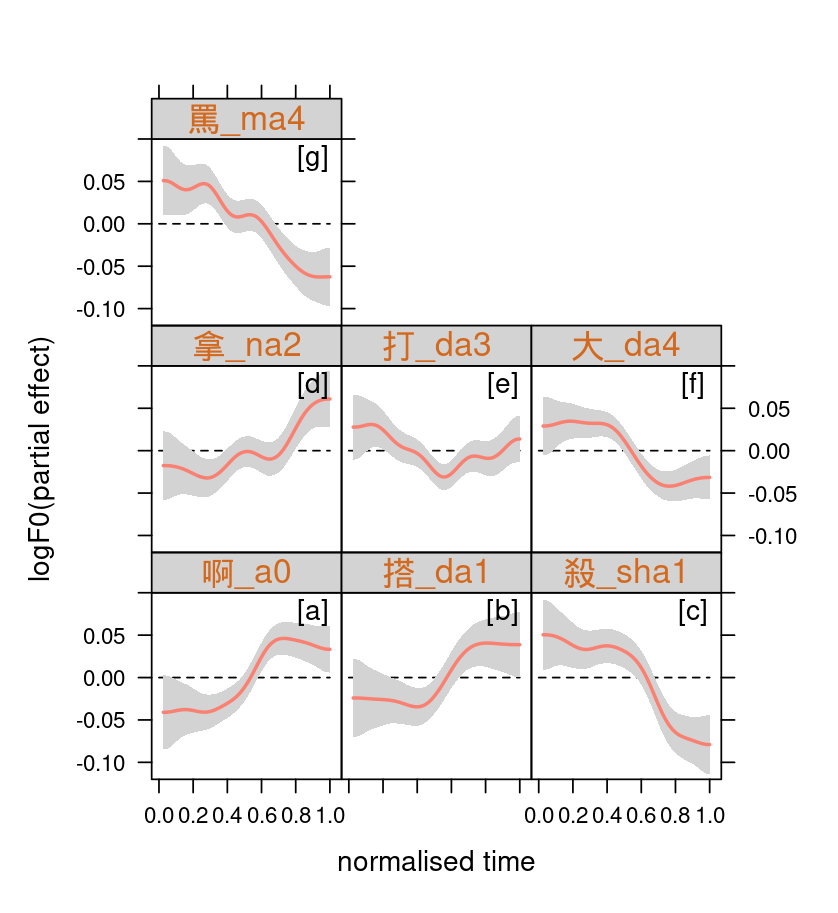}
    \end{minipage}
    \begin{minipage}[]{0.48\textwidth}
    \centering
        \includegraphics[width=1\textwidth]{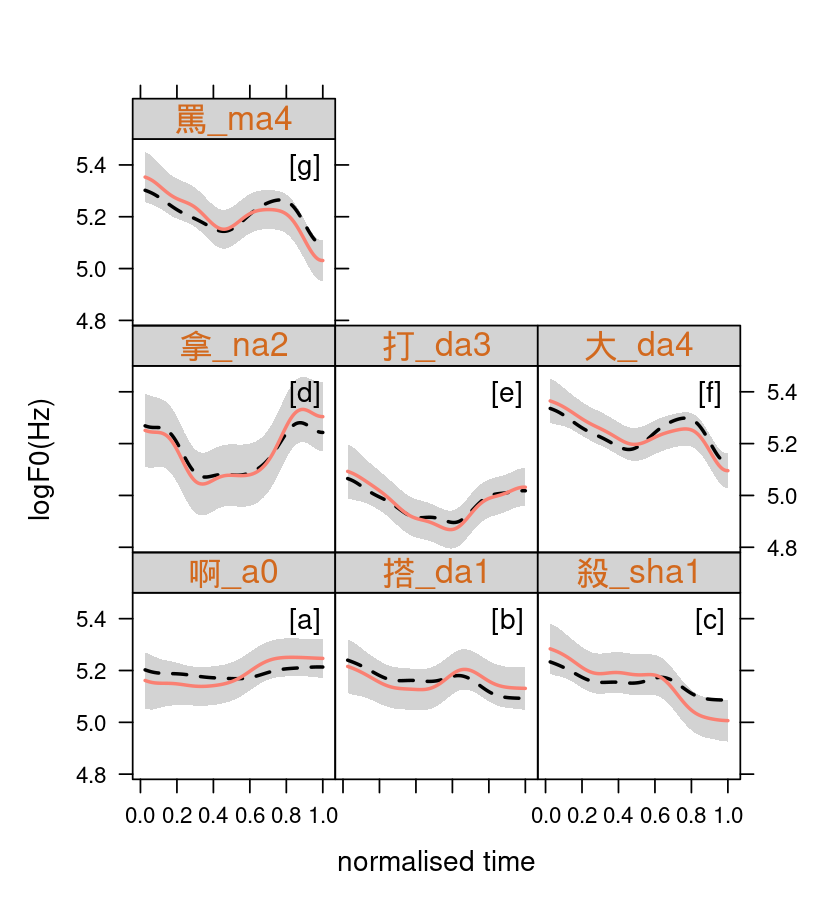}
    \end{minipage}
    \caption{Examples of word-specific modulation of pitch contours for words with /a/. Left panel: partial effect of word. Right panel: the predicted contour without the effect of word (black dashed line), and the predicted contour with the partial effect of word included (orange line with confidence interval) for female speakers. Predictions are for median \textit{duration},  \textit{semantic relevance} and \textit{speaker} set to the first speaker in the dataset. Utterance position is set to 1,  with the adjacent tone pattern fixed at `NULLNULL' (preceding and following pause).} 
    \label{fig:word}
\end{figure}

The left panel of Figure~\ref{fig:word} presents a selection of word-specific partial tone contours for words with the vowel /a/. Panel b and c present partial effects for three words that show that the by-word smooth does not simply replace the effects of \textit{tone pattern} (which is not taken into consideration in this GAM): these by-word partial effects deviate from the expected effects of \textit{tone pattern}.  The partial effects in panels (d--g) are somewhat more in line with their canonical tone contours.  The effect size of all by-word smooths is relatively small, well within the (-0.1, 0.1) range.  

The right panel clarifies how these word-specific partial effect smooths modulate the predicted pitch contours for female speakers, also taking into account all predictors of the baseline model.  To facilitate comparisons, we show the predicted contours, with (orange line) and without (black dashed line) the by-word effect, for when the word follows and precedes pauses (and \textit{utterance position} is necessarily 1). We set \textit{speaker} to the first speaker in our dataset, and we set word \textit{duration} and \textit{semantic relevance} to their median values.  As the by-word partial effects are relatively small, it is unsurprising that the extent to which they modulate the overall tone pattern, which to a large extent is determined by \textit{tone sequence}, is modest. 
\end{CJK}

\begin{CJK}{UTF8}{bsmi}
For the sentence-final modal particle 啊 (a0, right panel a), the contour without the word effect is level, the partial effect of \textit{word} changes it into a slightly rising tone.  For 搭 (d\={a}, `take', right panel b), the word effect rotates the effect of the other predictors, lowering pitch early in the word, and increasing it later in the word. The result is an overall reduction of the downward trend.  For 殺 (sh\={a}, `kill', right panel c), by contrast, the downward trend calculated from the other predictors is somewhat enhanced by the effect of \textit{word}.  In the case of  打 (d\v{a}, `hit', right panel e), the partial effect enhances the dipping tone already predicted by the other dependent variables. Finally, for  大 (d\`{a}, `big', right panel f), the word-specific effect enhances the fall that is only partially predicted by the other variables.

\subsubsection{Pitch contours for heterographic homophones}

We have seen that word co-determines the fine details of the pitch contours of monosyllabic words in Taiwan Mandarin, substantially outperforming \textit{tone pattern} and \textit{consonant} with respect to model fit. If it is indeed word meaning that is fine-tuning words' pitch contours, then heterographic homophones are expected to have their own pitch signatures.

\begin{figure}[H]
 \centering
\includegraphics[width=1\textwidth]{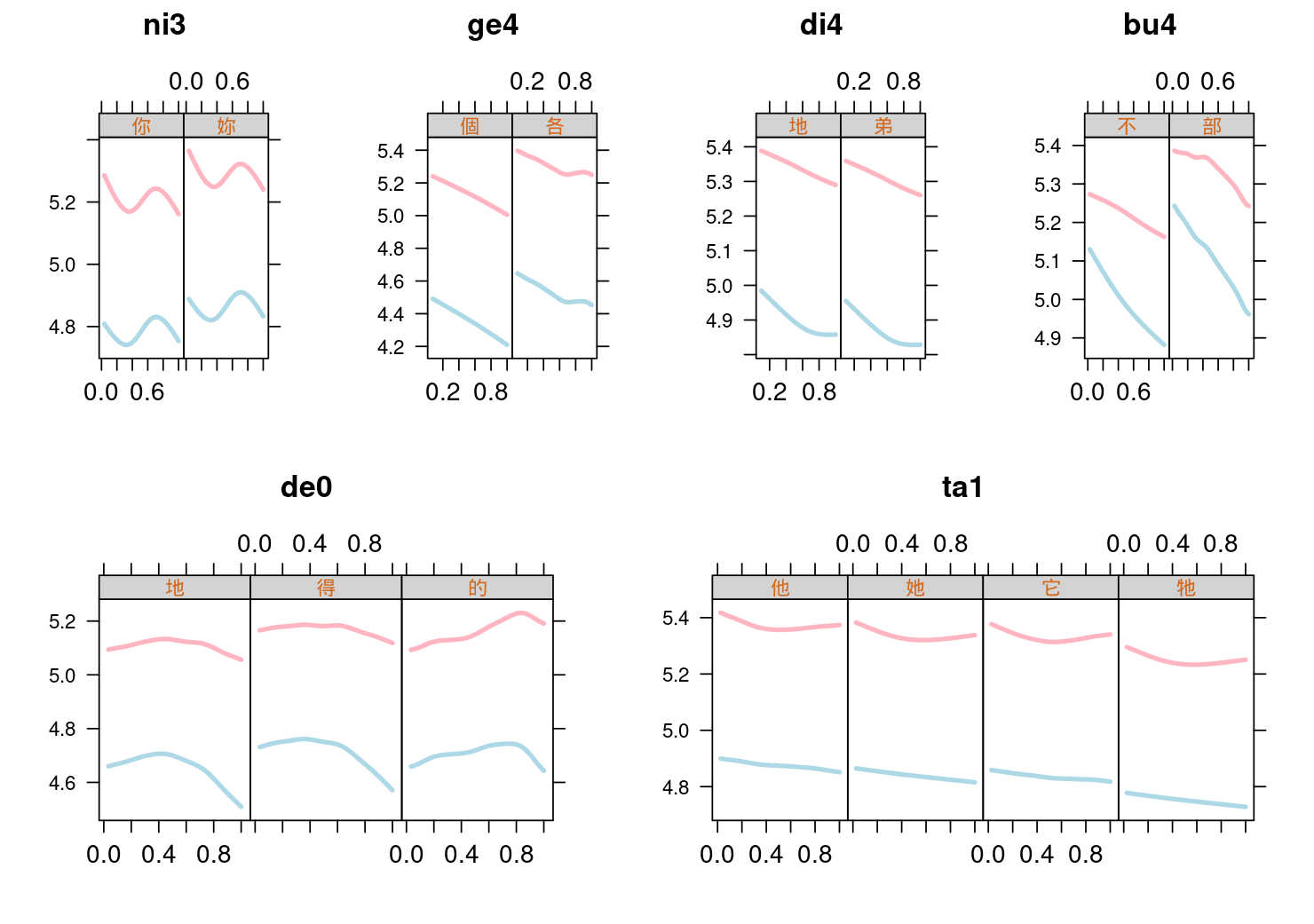}
\caption{The estimated pitch contours of heterographic homophones in our dataset. The horizontal axis represents normalised \textit{time}, and the vertical axis log-transformed F0.  Panels show the combined effect of intercept, \textit{gender} by \textit{time}, the effect of \textit{word}, and are calibrated for median \textit{duration}. Pitch is presented in pink for females, and in blue for males.}
\label{fig:homo}
\end{figure}

Our dataset contains six sets of heterographic homophones: four doublets (for ni2, ge4, di4, and bu4), a triplet (de0), and a quadruplet (ta1). We fitted a separate GAM to each of these six sets, including all the control variables from the baseline model, and a factor smooth for the homophones, as given by their characters. The resulting partial effects are visualized in Figure~\ref{fig:homo}.

The phonological word `ni3' is represented in Taiwan Mandarin by two characters,  妳 for female second person singular, and  你 for male second person singular. The pitch of   妳 has a higher onset than  你, for both male and female speakers.

Turning to  `ge4', we can see that the measure word  個 is realized with a slightly lower pitch than   各 (`each').  For  各, but not for  個, the fall levels off near the end of the word.  

The next panel presents the pitch contours for `di4' ( 地 `ground',  弟 `brother'). The kinship term is realized with an overall lower contour.  Male speakers realize these words with a fall that levels off.   For `bu4', for both female and male speakers, the fall has a stronger gradient for  部 (`department')  compared to  不 (`not').  

The lower left panel presents the pitch contours for three homophonic function words, all pronounced as /d\textschwa/.  The character  地, which when realized with T4 and with /i/ denotes `ground',  appears again in the lower left panel (de0), but now it is pronounced as /d\textschwa/.  This function word appears in constructions where it is preceded by a verb or adjective, and followed by another verb, and realizes a meaning similar to English adverbial \textit{-ly}. Its homophone  的 is used both as genitive marker and as complementiser. The third homophone,  得, is also an adverbial marker that appears in constructions where it is preceded by a verb or adjective, and followed by an adverb or adjective.  Even for native speakers, selecting which of the three characters to use is far from trivial, and in mainland China, spelling rules for less formal registers have recently been relaxed and allow the genitive marker  的 to be used across all constructions.  Nevertheless, there appear to be some differences in how these three words are pronounced.  The contours for men, showing a strong downward trend, contrast with those for women, whose contours are more level, or even show a rising trend.  For women, the genitive marker  的 has the greater rise compared to men, whereas for  得 and  地, the rise for women levels off more quickly compared to the men.

The lower right panel of Figure~\ref{fig:homo} concerns four heterographic homophonic personal pronouns, all of which share the same canonical high tone and denote third  person singulars:   她 (she),  他 (he),  它 (inanimate it) and   牠 (animate it) is specific to Taiwan Mandarin. In standard Mandarin, only  她,  他, and  它 are in use.  Women realize a slightly dipping tone, and men a falling tone on all four forms.  The difference in tonal realizations of the four pronouns is restricted to the intercept, which is maximal for  他, slightly lower for  她, and lowest for  牠.

\clearpage

\subsection{Sense-specific tonal contours}

The results of the preceding section support the possibility that word meaning is co-determining the way in which in Taiwan Mandarin pitch contours are realized. In this section, we consider whether we can replicate the results of \citet{Chuang:Bell:Tseng:Baayen:2024} and \citet{lu2024form}, who reported that model fit improved even further when by-word factor smooths for \textit{time} are replaced with by-sense factor smooths. Their finding suggests, as well as the results for heterographic homophones, strongly suggest that meaning is an important co-determinant of words' F0 contours.  

We therefore examined a GAM in which word sense replaced word. The senses of individual word tokens were estimated using the method described in \citet{Chuang:Bell:Tseng:Baayen:2024}, which makes use of the Chinese Wordnet and BERT-derived contextualized embeddings
\citep{huang2010constructing,hsieh2020tutorial}.  By way of example, this method assigns the sense `to read' to  讀 (d\'{u}) in  讀漢書 (d\'{u} h\`{a}n-sh\={u}, read History of the Former Han) or  讀史記 (d\'{u} sh\v{\i}-j\`{i}, read Records of the Grand Historian), but the sense `to study' when  讀 (d\'{u}) is used in the phrases  讀六年級 (d\'{u}-li\`{u}ni\'{a}nj\'{i}, study in sixth grade) or in  讀小學 (d\'{u}-xi\v{a}oxu\'{e}, study in elementary school). 

Figure~\ref{fig:AIC_F0} (presented earlier in section \ref{sec:canonical tone}) presents the improvement in AIC when word sense (red bars) is added to the baseline model.  For the words with /a/ and /\textschwa/, we observe a substantial improvement in model fit compared to the model with word. For /u/, the improvement is more modest, and for /i/, model fit is slightly inferior (by 6 AIC units).  Overall, \textit{word sense} emerges as the best predictor, outperforming \textit{tone pattern}, the initial \textit{consonant}, and \textit{word}.

\begin{figure}[H]
\centering
\includegraphics[width=1\textwidth]{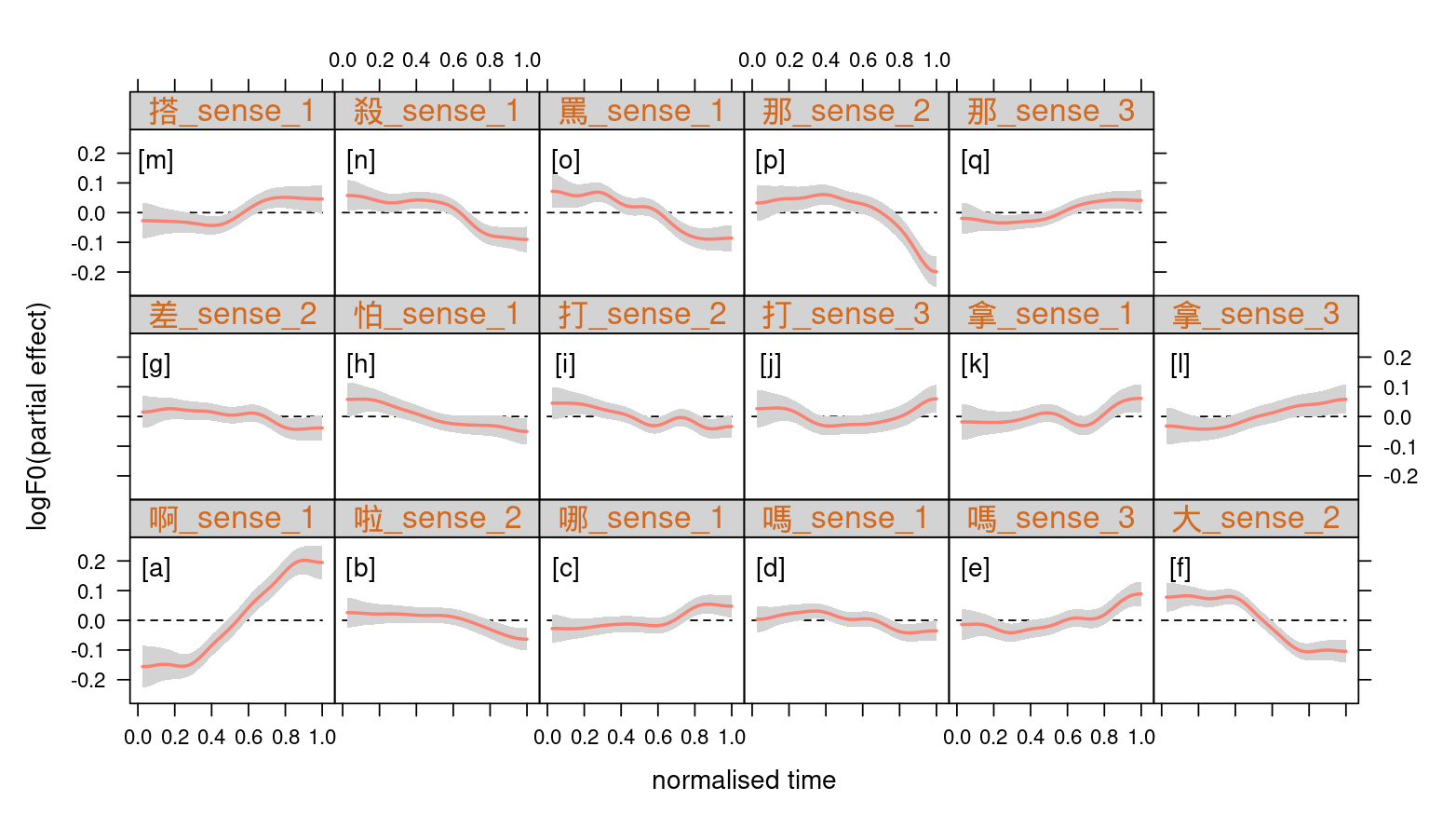}

\caption{Selected partial effects of word sense for words with the vowel /a/. For English translations, see Appendix~\ref{appendix translation}.}
\label{fig:sense_a}
\end{figure}

Figure~\ref{fig:sense_a} displays the partial effect for the subset of words with /a/ where clear sense-specific contours are present.  For some words (e.g.,  啊 (a0, sentence-final particle),  大 (d\`{a}, `big'),  差 (ch\={a}, `differ/error'), only one word sense is displayed, the other word sense or word senses having no well-supported distinct contours of their own.

The sentence-final particles  嗎 (ma0, panel d: pause marker, and panel e: to inquire about someone's wishes) show partial effects of falling and a rising contour, respectively.   打 (d\v{a}) in the sense of hitting something with hands or hand-held objects (panel i), as in the sentence  你家小孩做了你不喜歡的事情你就可以打他呢 (n\v{\i}-ji\={a}-xi\v{a}oh\'{a}i-zu\`{o}-le0-n\v{\i}-b\`{u}-x\={\i}hu\={a}n-de0-sh\`{i}q\'{i}ng-n\v{\i}-ji\`{u}-k\v{e}-y\v{\i}-d\v{a}-t\={a}-ne0, `When your kid did something you don't like, you can just hit him', sentence taken from the corpus), has a downward sloping partial effect pitch contour.  When  打 refers to a fight between two groups (panel j), it has a fall-rise partial effect contour,  as in  打架(d\v{a}ji\`{a}, `fight').  拿 is similar to English `take'.  It can mean using the hand to take or hold things (e.g., in the corpus  拿菜單, n\'{a}-c\`{a}id\={a}n, `take the menu'), but also `take from' (e.g.,  \begin{CJK}{UTF8}{gbsn}  拿钱 n\'{a}-qi\'{a}n, take money from somebody). With the former sense,  拿 is realized with a late rise, with the latter sense, the rise is initiated much earlier.  In expressions such as  那因为 (n\`{a}-y\={\i}nw\`{e}i4, `that's because') and  那是现在啦 (n\'{a}-sh\`{i}-xi\`{a}nz\`{a}i-la0, `that's nowadays' ),   那 (n\`{a}, panel p) is realized with a distinct falling partial effect pitch contour, but in deictic expressions such as  那时候 \end{CJK}(n\`{a}-sh\'{i}ho\`{u}, `at that time') and  那個親戚 (n\`{a}-g\`{e}-q\={\i}nq\`{i}, `that relative'),  it is realized with a slightly rising partial effect pitch contour(panel q).  Finally, we note that although the modal particles  哪 (na0) and  啊 (a0) are assigned the same sense, `explaining' or `reminding',  啊 (panel a) has a significant rising partial effect tone contour compared to the nearly flat partial effect pitch contour that characterizes  哪 (panel c). Possibly, the sense disambiguation algorithm failed to tease the senses of these sentence-final particles apart ---  the subtle nuances in meaning of sentence-final particles are difficult to articulate even for native speakers.  This finding supports the distinction introduced above in section~\ref{sec:word sense} between \textit{sense} and \textit{word sense}, a distinction that is considered further in the random forest analysis reported below in the next section. 

\subsection{Additional statistical analyses}\label{sec: gam_randomforest}

We carried out two further statistical analyses to consolidate the results reported in the preceding sections. The first analysis again makes use of a GAM, but the model is now fitted to the data containing all vowels. It has \textit{word sense} because it appeared to outperform \textit{tone pattern}, \textit{consonant}, and \textit{word}. This model included the main effect of vowel in order to clarify whether high vowels are indeed produced with higher pitch compared to low vowels. We refrained from including interactions with vowel. Given the models for the individual vowels reported above, any differences are expected to be small to contribute little to the model fit, making the model unnecessarily complex while increasing the risk of overfitting. The second analysis uses a random forest as an independent method for assessing the relative importance of the different predictors.

Table~\ref{tab.gam} provides the summary of the GAM fitted to the data of all four vowels jointly, with by-\textit{word sense} factor smooths for normalised \textit{time}, the control variables of the baseline model that we also took into account when modeling the data of the individual vowels, and also included \textit{tone pattern}. As the effect of \textit{tone pattern} that emerged from the preceding analysis mainly concerns changes in pitch height, but not in pitch contours, we did not include in the present omnibus model a smooth for the interaction of \textit{tone pattern} by normalized \textit{time}.  In the resulting model, the effect of \textit{tone pattern} is captured by adjustments to the intercept (see the upper part of Table~\ref{tab.gam}; the neutral tone is the reference level, the estimated coefficients represent treatment contrasts).  

\begin{table}[ht]
\centering
\begin{tabular}{lrrrr}
   \hline
\textbf{A. parametric coefficients} & \textbf{Estimate} & Std. Error & t-value & p-value \\ 
  Intercept & 5.1465 & 0.0282 & 182.7152 & $<$ 0.0001 \\ 
  gender:  男 & -0.4091 & 0.0337 & -12.1481 & $<$ 0.0001 \\ 
  tone\_pattern: T1 & 0.0991 & 0.0236 & 4.1948 & $<$ 0.0001 \\ 
  tone\_pattern: T2 & -0.0576 & 0.0289 & -1.9894 & 0.0467 \\ 
  tone\_pattern: T3 & -0.0640 & 0.0234 & -2.7360 & 0.0062 \\ 
  tone\_pattern: T4 & 0.1608 & 0.0235 & 6.8419 & $<$ 0.0001 \\ 
  vowel: /\textschwa/ & 0.0235 & 0.0164 & 1.4292 & 0.1529 \\ 
  vowel: /i/ & 0.0727 & 0.0176 & 4.1195 & $<$ 0.0001 \\ 
  vowel: /u/ & 0.0188 & 0.0204 & 0.9214 & 0.3569 \\ 
   \hline
\textbf{B. smooth terms} & edf & Ref.df & F-value & p-value \\ 
  s(normalised time): gender  女 & 2.9110 & 2.9555 & 39.0680 & $<$ 0.0001 \\ 
  s(normalised time): gender  男 & 2.8653 & 2.9382 & 17.0272 & $<$ 0.0001 \\ 
  s(log\_duration): gender  女 & 2.7104 & 2.9291 & 9.8467 & 0.0001 \\ 
  s(log\_duration): gender  男 & 2.6963 & 2.9266 & 6.2416 & 0.0004 \\ 
  ti(normalised time, log\_duration) & 7.2067 & 7.9494 & 27.8391 & $<$ 0.0001 \\ 
  s(normalised time, word sense) & 653.3787 & 1015.0000 & 4.3264 & $<$ 0.0001 \\ 
  s(speaker) & 52.1480 & 54.0000 & 93.8580 & $<$ 0.0001 \\ 
  s(semantic relevance) & 2.8860 & 2.9887 & 8.4289 & $<$ 0.0001 \\ 
  s(normalised time, tone sequence) & 765.8064 & 1554.0000 & 2.3003 & $<$ 0.0001 \\ 
  s(utterance position) & 2.1437 & 2.5041 & 130.6635 & $<$ 0.0001 \\ 
   \hline
\end{tabular}
\caption{Summary of a GAM fitted to the data of all vowels.  The reference level for \textit{gender} is female, the reference level for tone is the neutral tone, and the reference level for vowel is /a/. As the residuals of this model do not show a Gaussian distribution, p-values are approximations only. The auto-correlation parameter was set to $\rho = 0.9$.} 
\label{tab.gam}
\end{table}

This omnibus GAM replicates the results from the separate GAM models for the different vowels above. In addition, it shows differences between the vowels. Words with the high vowel /i/ have a somewhat higher pitch than words with other vowels, which dovetails well with the cross-linguistic observation that high vowels tend to be realized with high tones \citep{ho1976acoustic,bo1987vowel}. However, for /u/, which also is a high vowel, this generalization does not hold.

In order to rule out the unlikely possibility that the greater importance of \textit{word} and \textit{word sense} as compared to \textit{tone pattern} and initial \textit{consonant} is an artifact of the generalized additive model, we investigated the variable importance of our predictors also using random forests \citep{breiman2001random} as implemented in the \textbf{party} package for R \citep[cf][]{strobl2009introduction}. 
A question that arises at this point is whether the random forest should predict the observed pitch, or instead pitch centered for each individual token. In the latter analysis, the random forest is forced to look more closely at developments over normalised time. In the former analysis, \textit{speaker} will dominate the model, as the differences in average pitch between individual speakers are large.  In what follows, we report the random forests for the uncentered, log-transformed pitch contours, in order to give the canonical tones patterns (which according to the GAM analyses have different average pitch) maximal opportunity to contribute to prediction. However, random forests predicting centered pitch produce results that also support the conclusions reported below.

In the random forest analysis, we considered two sense-related variables, the first of which, \textit{word sense}, was used in the preceding analyses. \textit{Word sense} assumes that words that are written with different characters always have different senses.   There are some words that are written differently, but share the same sense. For instance, the sense `expressing end of turn' is shared by  啊,  哪,  吧,  嗎 and  啦. \textit{word sense} denotes the combination of a word's character and its sense. Thus, the variable labelled \textit{word sense} labels  啊,  哪,  吧,  嗎 and  啦 as five distinct senses, instead of one.  The variable \textit{sense} treats shared senses of words written with different characters as the same sense. We included this additional predictor in order to assess whether the specific way in which sense is defined might lead to different conclusions. 

\begin{figure}[H]
    \centering
    \includegraphics[width=1\textwidth]{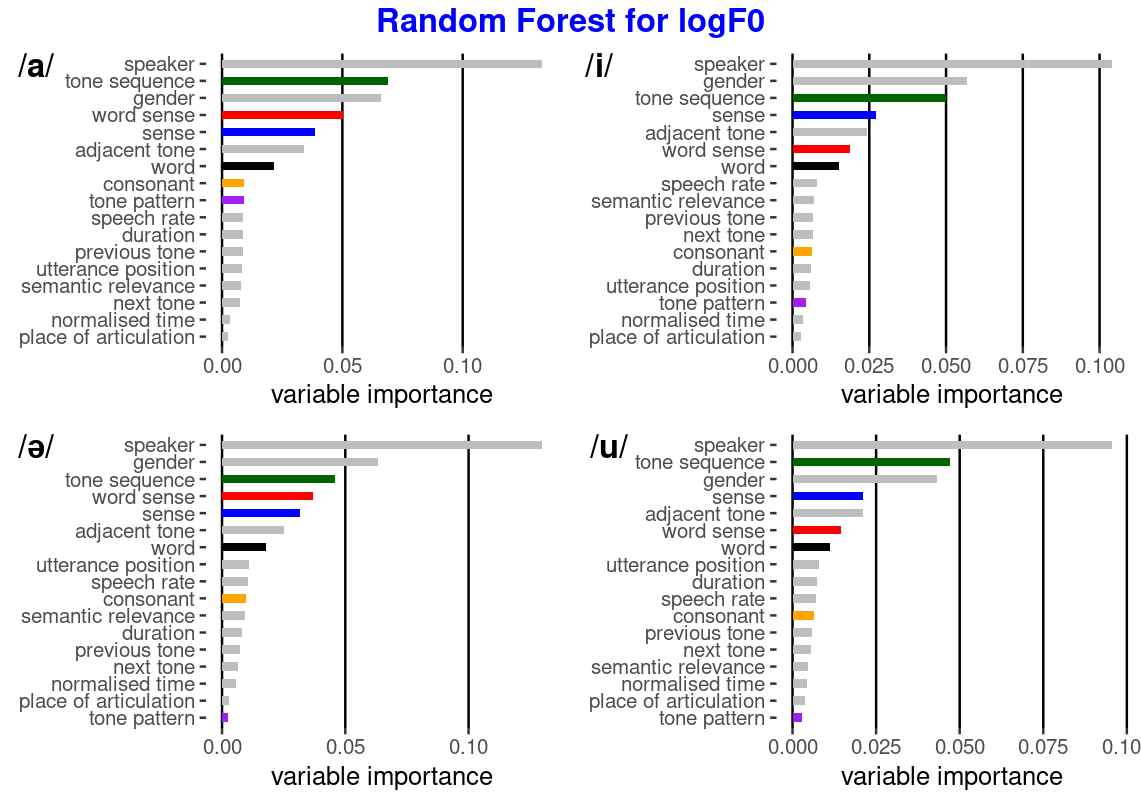}
    \caption{Variable importance for the predictors in random forest analyses with log-transformed pitch as response variable for four vowels. Core predictors (\textit{word sense} or \textit{sense}, \textit{word}, \textit{consonant} and \textit{tone pattern}) and \textit{tone sequence} are marked in different colours.}
    \label{fig:random forest F0}
\end{figure}

As can be seen in Figure~\ref{fig:random forest F0},  \textit{tone sequence} ranks among the top three most important predictors, together with \textit{speaker} and \textit{gender}.  \textit{Word sense} (red) or \textit{sense} (blue), also have high variable importance, exceeding those of \textit{word} (black), \textit{consonant} (orange), and \textit{tone pattern} (purple), across all vowels. This also hold for the vowel /i/, even though, for this vowel, the GAM model did not show that \textit{sense} was a more important predictor than \textit{word}.  Furthermore, \textit{word} also outperforms \textit{consonant} and \textit{tone pattern} for each of the four vowels.   Reassuringly, the two sense variables have very similar variable importances across the vowels. A figure for the variable importances of the predictors in a random forest for the data of all vowels jointly is available in the appendix~\ref{appendix_rf}, and provides a very similar evaluation of variable importance. The fact that vowel has the lowest variable importance of all predictors justifies our decision not to include interactions with vowel in the omnibus GAM reported above.

To summarize, considered jointly, the results of both the GAM models and those of the random forests support the importance of word and word sense as co-determinants of words' tone contours, showing that the findings of \citet{Chuang:Bell:Tseng:Baayen:2024} and \citet{lu2024form} are not restricted to disyllabic words.  Furthermore, both the GAMs and the random forests point to an important role of the tonal context in which a word appears.  By contrast, the effects of \textit{consonant}, \textit{vowel}, and \textit{tone pattern} are considerably less well supported.

\section{General discussion} \label{sec:gendisc}

\noindent
This study investigated the F0 contours of monosyllabic words in spontaneous conversations in Taiwan Mandarin.  Our first finding is that the tonal context (represented by the predictor \textit{tone sequence}) has a very strong effect on the realization of pitch contours.  The effect of tonal sequence is much stronger than that of the canonical tone patterns as described in the literature and textbooks.  

A closely related finding is that once the effect of \textit{tone sequence} is taken into account, the tones do not all surface as expected given the literature. T1, T2 and T3 all emerged as level tones, with T1 a high tone (as expected), and T2 and T3 as low tones, although these two tones are assumed to be rising and dipping, respectively  \citep[for the realization of T3 as a low level tone in Taiwan Mandarin, see also][]{fon2004production,fon2007positional}.  T4 emerged as a falling tone starting high, as expected, but with a highly restricted downward pitch movement.   Even in words surrounded by pauses, the canonical tone contours do not emerge in a consistent way. Figure~\ref{fig:adj-tone}, for instance, suggests that between pauses, as expected, T1 is a high level tone and T3 a dipping tone, but T2 also appears to be a dipping tone, and T4 a tone with a fall-rise-fall contour.

We cannot explain the observed effects of tonal context from articulatory constraints, for two reasons.  First, there is enormous variation in the effects of neighboring tones, with some effects aligning with the theoretical literature, and others diverging.  Second, when we started this investigation, we assumed that the tone contours of neighboring words are adequately described by their canonical tones.  However, this is not the case. As a consequence,  the canonical tones from adjacent words constitute a sub-optimal starting point for inferences about tonal co-articulation.  For future research, working with properties of the actual pitch in adjacent words is recommended.

The strong effects of \textit{tone sequence} contrast with the laboratory speech investigated by, e.g., \cite{xu1994production,li2016acoustic,liu2022context}. In these  earlier studies, the effect of tonal context was either not taken into account (when considering words spoken in isolation) or taken into account only to a limited extent (when considering words spoken in a few controlled sentence contexts).  However, our results demonstrate that the wide variety of tonal contexts in which words occur are an important source of substantial tonal variation. One possible reason for the strong effect of tonal context in our data is that, in careful, deliberate speech, speakers tend to enunciate more clearly, reducing the effects of co-articulation. In comparison, in spontaneous speech, where speakers are in a more natural and informal atmosphere, assimilation effects are stronger. 

A second important finding is that the tonal realization of monosyllabic words is partly determined by their meanings.  The senses of monosyllabic Mandarin tokens (determined on the basis of the context in which they occur) enable the GAM to model pitch contours more precisely than is possible with words' canonical tones.  \textit{Word sense} as a predictor also outperforms \textit{word}, although \textit{word} by itself already outperforms canonical \textit{tone pattern} by far.  These findings are consistent with the results reported for disyllabic words with the T2-T4 tone pattern by \citet{Chuang:Bell:Tseng:Baayen:2024} and disyllabic words with the T2-T3 and T3-T3 tone patterns by \citet{lu2024form}.

Third, in line with this finding, we also observed that heterographic homophones in Mandarin tend to have distinct F0 contours. This observation provides further support for the hypothesis that words' meanings co-determine how pitch contours are realized. 
 
Fourth, effects of the segmental properties were small, both according to the GAM models and according to the random forests. We see differences between words with different vowels, although the effects are small. The pitch for words with /i/ is generally higher than for words with the other vowels, in accordance with the literature reporting that high vowels favor high tones.  However, we did not observe the same pattern for /u/.  

Further, like \cite{xu2003effects},  we observed a difference between words starting with aspirated and non-aspirated stops. However, whereas \cite{xu2003effects}  reported higher pitch onset for words with /t/ compared to /t\super h/, in our corpus, differences in pitch varied inconsistently across different pairs of aspirated and unaspirated stops. Nevertheless, the initial \textit{consonant} is an important predictor of F0, and a better predictor than canonical \textit{tone pattern}.  This may be because the initial consonant provides information about the identity of the word and of its meaning. This hypothesis is supported by the predictors \textit{word} and \textit{word sense} , which have larger effects than \textit{consonant}.

Fifth, although the neutral tone is described in the literature as being completely dependent for its realization on the preceding tone \citep{chao1930system,yip1980tonal,chien2021investigating}, our investigations suggest that the neutral tone is a mid-to-low tone that is subject to the same variable effects of surrounding tones as the T1 to T4.  In other words, the neutral tone is a tone in its own right that is shaped by the same forces that also co-determine the details of the realization of the standard tones T1, T2, T3, and T4.


In Mandarin textbooks, descriptions of the tonal system of Mandarin Chinese take the canonical lexical tones as givens \citep{xinhuadictionary}. While this may be correct for careful speech, in casual speech, at least in Taiwan Mandarin, and we suspect also in other dialects of Mandarin, actual tonal realizations diverge considerably from what the textbooks would lead one to expect.  We anticipate that also in the higher registers of spoken Mandarin Chinese, tonal co-articulation is present, and that tone is also co-determined by words' senses.  If future research uncovers further support for this conjecture, then this will help explain why second language learners of Mandarin Chinese, being presented with relatively careful speech, often struggle with learning the tones \citep{hao2012second}: the information on tone gleaned from pinyin transcriptions and the actual properties of the F0 contour all too often diverge.

Our findings concerning strong tonal co-articulation and additional word-sense specific tonal realization may also help explain why Deep Learning is particularly successful in speech technology \citep{chen2016tone,tang2021end}. By learning from vast amount of speech data, these models become sensitive to the nuanced variation in contours, allowing them to better understand and produce spontaneous speech.

In conclusion, our study offers a comprehensive exploration of the realization of monosyllabic words in a corpus of spontaneous Taiwan Mandarin, leveraging the power of the generalized additive model to understand the way in which a wide range of factors, including tonal co-articulation and word meaning, work together to shape F0 contours.   The new perspective on the realization of pitch is both more complex and more rich than what standard analyses suggest. However, we suspect that with our study we have just scratched the surface of the intricacies of Mandarin tone.

\clearpage
\appendix
\section*{Appendix}
\setcounter{figure}{0}

\section{Time by duration by gender for four vowels}\label{appendix duration}

The effect of \textit{duration} varies significantly among the four different vowels and across the two genders. For female speakers, pitch contours remain relatively flat in vowels of short and medium duration. For longer words with /i/, a fall-rise inflection is present that becomes more pronounced with increasing duration. For words with /u/, long durations are paired with a strong downward curvature that  levels off near the center of the vowel and then shows a slight increase.   For words with /a/, as duration increases, the intercepts of pitch increase, going hand in hand with increasingly steep falls.  Finally, a fall-rise inflection is present for /\textschwa/ for longer durations, in combination with a general lowering of pitch. Somewhat similar results are visible for male speakers. However, for them, for words with /a/, a higher duration goes hand in hand with a somewhat lower pitch, contrasting with the increasing pitch for female speakers. For words with the vowel /\textschwa/, short durations show a downward trend, which morphs into a U-shaped trend for long durations. The omnibus GAM for all vowels includes well-supported  interactions of \textit{time} by \textit{gender}, \textit{duration} by \textit{gender}, and \textit{time} by \textit{duration} (see Table~\ref{tab.gam}). It is possible that a three-way interaction of \textit{time} by \textit{duration} by vowel is present, but we leave exploration of this complex interaction for further research.

\renewcommand{\thefigure}{A\arabic{figure}}
\begin{figure}[H]
\centering
\includegraphics[width=1.0\textwidth]{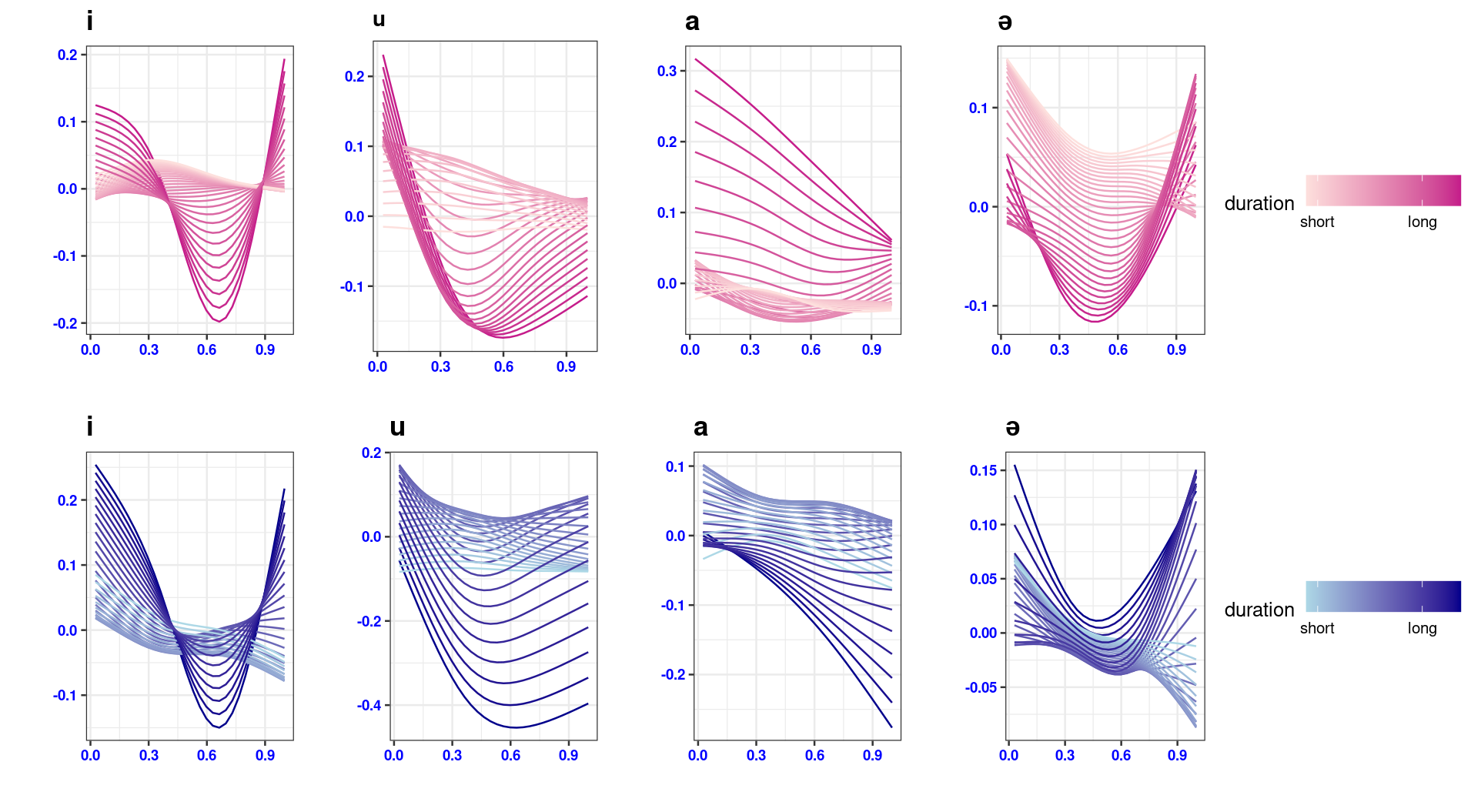}
\centering
\caption{Upper panel: the partial effect of \textit{duration} for female speakers; Lower panel: the partial effect of \textit{duration} for male speakers. The range of the log-transformed of pitch on the y-axis on both panels differs according to the corresponding vowel.}
\label{fig:duration_4)}
\end{figure}

\newpage

\section{Word senses distinguished for 13 characters}\label{appendix translation}
\setcounter{table}{0}
\renewcommand{\thefigure}{A\arabic{figure}}
Complementing Figure~\ref{fig:sense_a}, the following table lists the senses and their definitions in Mandarin and English. \vspace*{2\baselineskip}
\nopagebreak
\footnotesize{
\begin{table}[H]
    \centering
    \resizebox{1.0\linewidth}{!}{
    \begin{tabular}{lp{2cm}p{7cm}p{7cm}p{2cm}}
    \hline
         Character&Sense & Sense Meaning & Translation\\\hline
          啊 & sense\_1 &  表解釋或提醒對方的語氣 & indicate a tone used to explain or remind someone\\
          啦 & sense\_2 &  表停頓的語氣 & indicate a pause in speech \\
          哪 & sense\_1 &  表解釋或提醒對方的語氣 & indicate a tone used to explain or remind someone\\
          嗎 & sense\_1 &  表停頓的語氣 & indicate a pause in speech\\
          嗎 & sense\_3 &  表徵詢對方意願的疑問語氣 & to ask about others' preference\\
          大 & sense\_2 &  形容體積超過比較對象的 & describe something as larger in size than the object of comparison (big)\\
          差 & sense\_2 &  形容對特定對象的評價是負面的 & a negative evaluation of a specific object \\
          怕 & sense\_1 &  認为後述事件可能發生而擔心产生負面的結果 & express fear of a potential future event that might lead to negative consequences \\
          打 & sense\_2 &  兩對立團體之間起武力衝突 & a physical conflict between two opposing groups (fight)\\
          打 & sense\_3 &  用手或手持物打後述對象,使其感到疼痛或受到傷害 & to strike a person with a hand or object, causing pain or injury (hit) \\
          拿 & sense\_1 &  用手取物或持物 & to take or hold something with one's hands\\
          拿 & sense\_3 &  向特定對象取得主事者擁有或應該擁有的金錢或物品 & to obtain money or items from someone that the owner or rightful person possesses\\
          搭 & sense\_1 &  搭乘有座位的交通工具 & to take a seat on a mode of transportation\\
          殺 & sense\_1 &  使動物失去生命 & to cause an animal to die (kill)\\
          罵 & sense\_1 &  以粗話惡語侮辱人 & to insult someone with vulgar language (scold)\\
          那 & sense\_2 &  代指那時,有加强語氣的作用 & refer to a specific moment in time, with emphasis\\
          那 & sense\_3 &  根據说話者说話時所處的時空而言,指稱比較遠的後述對象 & refer to something distant in space or time, relative to the speaker (that)\\\hline
    \end{tabular}}
    \caption{Senses referenced in Figure~\ref{fig:sense_a}.}
    \label{tab:sense_translation}
\end{table}
}
\newpage

\section{Variable importance according to a random forest for the data of the four vowels jointly}\label{appendix_rf}
Variable importance according to a random forest of predictors for the uncentered, log-transformed pitch contours, conducted on the dataset containing all four vowels. Vowel (indicated by the green bar) has the lowest impact on the pitch contours of monosyllabic Mandarin words. Similar to the results shown in Figure~\ref{fig:random forest F0}, \textit{tone sequence} and \textit{word sense} are the two most crucial predictors, followed by \textit{word}, \textit{consonant}, and \textit{tone pattern}.

\begin{figure}[H]
    \centering
    \includegraphics[width=0.8\linewidth]{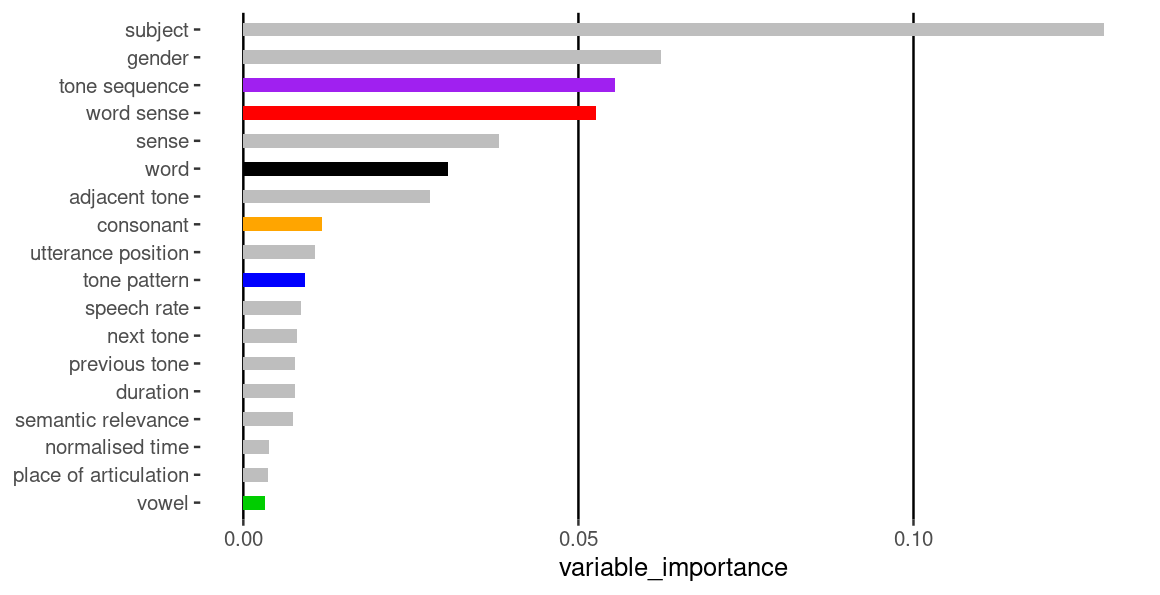}
    \caption{Variable importance for the predictors in random forest analyses for the data of all vowels jointly}
    \label{fig:random_forest_full}
\end{figure}

\clearpage
\bibliography{references}

\end{CJK}
\end{CJK}
\end{document}